\documentclass[10pt]{article} 
\PassOptionsToPackage{dvipsnames}{xcolor}
\usepackage[accepted]{rlj} 

\usepackage{amssymb}            
\usepackage{mathtools}          
\usepackage{mathrsfs}           
\usepackage{graphicx}           
\usepackage{subcaption}         
\usepackage[space]{grffile}     
\usepackage{url}                
\usepackage{lipsum}             
\usepackage{amsthm}

\usepackage{algorithm}
\usepackage{algpseudocode}
\usepackage{wrapfig}
\usepackage{booktabs}
\usepackage{multirow} 
\usepackage{listings}
\usepackage{footnote}

\definecolor{codebg}{rgb}{0.93, 0.93, 0.93}
\algdef{SE}[SUBALG]{Indent}{EndIndent}{}{\algorithmicend\ }%
\algtext*{Indent}
\algtext*{EndIndent}

\newtheorem*{theoremfarahmand*}{Theorem 3.4 from \citet{farahmand2011regularization}}

\title{Gradient Iterated Temporal-Difference Learning}
\setrunningtitle{Gradient Iterated Temporal-Difference Learning}

\author{
Théo Vincent\textsuperscript{1,2, $\dagger$}, Kevin Gerhardt\textsuperscript{1,2}, Yogesh Tripathi\textsuperscript{1,2}, Habib Maraqten\textsuperscript{1,2}, \\ Adam White\textsuperscript{3,4,5}, Martha White\textsuperscript{3,4,5}, Jan Peters\textsuperscript{1,2,6,7}, Carlo D'Eramo\textsuperscript{8}
}

\emails{\vspace{-0.1cm}}

\affiliations{
$^{1}$\textbf{DFKI, SAIROL}, 
$^{2}$\textbf{TU Darmstadt},
$^{3}$\textbf{Alberta Machine Intelligence Institute (Amii)},
$^{4}$\textbf{Department of Computing Science, University of Alberta, Canada},
$^{5}$\textbf{CIFAR AI Chair},
$^{6}$\textbf{Hessian.ai},
$^{7}$\textbf{Robotics Institute Germany},
$^{8}$\textbf{University of Würzburg} \\
$^\dagger$ correspondence to \url{theovincentjourdat@gmail.com} 
}

\contribution{
    We introduce \textit{Gradient Iterated Temporal-Difference} learning, a new gradient temporal-difference learning algorithm, which is designed to learn a sequence of action-value functions in parallel, where each function is optimized to represent the application of the Bellman operator over the previous function in the sequence (Section~\ref{S:gitd}).
    }
    {
    \citet{schmitt2022chaining} first introduced the idea of learning a sequence of action-value functions where each function represents the application of the Bellman operator over the previous function. They suggest performing updates sequentially, starting with the first function in the sequence. However, this method is not scalable as it relies on multiple sequential gradient steps. Then, \citet{vincentiterated} suggested performing the updates in parallel, but rely on semi-gradient updates, which can cause instability because each function is optimized with respect to a moving target. Finally, \citet{vincent2026bridging} developed a parameter-efficient version of this idea, which still relies on semi-gradient updates. Our method leverages the gradient TD method published by \citet{ghiassian2020gradient}.
    }

\contribution{
    We derive and evaluate three instances of the introduced algorithm to showcase that the presented method can be used in many different settings and combined with many deep reinforcement learning algorithms (Section~\ref{S:experiments}).
    }
    {
    We present combinations of the proposed algorithm with DQN, SAC, and CQL, along with an advanced architecture (IMPOOLA, \citet{trumpp2025impoola}), a 3-step return, and a prioritized replay buffer.
    }

\contribution{
    We demonstrate that the learning speed of the presented gradient TD method is competitive against semi-gradient methods across multiple benchmarks, including Atari games. In particular, no prior work has demonstrated that gradient TD methods are competitive in the Atari benchmark (Section~\ref{S:benchmark_performance}). 
    }
    {
    The proposed approach is competitive against semi-gradient methods such as DQN, SAC, and CQL, as well as their iterated variants, across $10$ Atari games and $6$ MuJoCo environments. \citet{ghiassian2020gradient, elelimydeep} evaluate their proposed gradient TD method on MinAtar~\citep{young19minatar}. While \citet{elsayed2024streaming}, \citet{vasan2024avg}, and \citet{jiang2022learning} report experiments on Atari games, these methods use eligibility traces or emphatic weightings, which rely on semi-gradient updates. 
    }


\summary{
\vspace{-0.25cm}
Temporal-difference (TD) learning is highly effective at controlling and evaluating an agent's long-term outcomes. Most approaches in this paradigm implement a semi-gradient update to boost the learning speed, which consists of ignoring the gradient of the bootstrapped estimate. While popular, this type of update is prone to divergence, as Baird's counterexample illustrates. Gradient TD methods were introduced to overcome this issue, but have not been widely used, potentially due to issues with learning speed compared to semi-gradient methods. Recently, iterated TD learning was developed to increase the learning speed of TD methods. For that, it learns a sequence of action-value functions in parallel, where each function is optimized to represent the application of the Bellman operator over the previous function in the sequence. While promising, this algorithm can be unstable due to its semi-gradient nature, as each function tracks a moving target. In this work, we modify iterated TD learning by computing the gradients over those moving targets, aiming to build a powerful gradient TD method that competes with semi-gradient methods. Our evaluation reveals that this algorithm, called \textit{Gradient Iterated Temporal-Difference} learning, has a competitive learning speed against semi-gradient methods across various benchmarks, including Atari games, a result that no prior work on gradient TD methods has demonstrated. \\
\centerline{\raisebox{-.15\height}{\includegraphics[height=0.75\baselineskip]{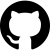}} ~\url{https://github.com/theovincent/Gi-DQN}~ \raisebox{-.15\height}{\includegraphics[height=0.75\baselineskip]{figures/github_logo.png}}}
\vspace{-0.5cm}
}

\begin{document}

\makeCover
\maketitle

\vspace{-0.7cm} 
\begin{abstract}
\vspace{-0.1cm} 
Temporal-difference (TD) learning is highly effective at controlling and evaluating an agent's long-term outcomes. Most approaches in this paradigm implement a semi-gradient update to boost the learning speed, which consists of ignoring the gradient of the bootstrapped estimate. While popular, this type of update is prone to divergence, as Baird's counterexample illustrates. Gradient TD methods were introduced to overcome this issue, but have not been widely used, potentially due to issues with learning speed compared to semi-gradient methods. Recently, iterated TD learning was developed to increase the learning speed of TD methods. For that, it learns a sequence of action-value functions in parallel, where each function is optimized to represent the application of the Bellman operator over the previous function in the sequence. While promising, this algorithm can be unstable due to its semi-gradient nature, as each function tracks a moving target. In this work, we modify iterated TD learning by computing the gradients over those moving targets, aiming to build a powerful gradient TD method that competes with semi-gradient methods. Our evaluation reveals that this algorithm, called \textit{Gradient Iterated Temporal-Difference} learning, has a competitive learning speed against semi-gradient methods across various benchmarks, including Atari games, a result that no prior work on gradient TD methods has demonstrated.
\vspace{-0.5cm} 
\end{abstract}

\section{Introduction}
\vspace{-0.2cm} 
Temporal-difference (TD, \citet{sutton1988learning}) learning is a useful learning paradigm that eliminates the need for long rollouts in the environment, thereby increasing the frequency with which the learning agent receives feedback. This paradigm is successful in solving complex tasks where sample efficiency is key \citep{mnih2015human, degrave2022magnetic, wurman2022outracing}. The core idea behind temporal-difference learning is to reduce the error between the estimated outcome at the current timestep and the target, where the target is the received reward plus the expected discounted outcome at the next timestep.

Applying stochastic gradient descent on this error is a straightforward approach. Unfortunately, this idea is limited to deterministic environments because two independent samples, each starting at the same timestep, are required to compute an unbiased estimate of the target. This issue is known as the double sampling problem~\citep{baird1995residual}. For this reason, semi-gradient methods, like TD(0), only use the gradient at the current timestep and ignore the gradient of the target~\citep{watkins1992q}. Interestingly, despite known divergence issues on simple problems due to the absence of this gradient term~\citep{baird1995residual}, this approach is pursued by all state-of-the-art TD learning methods~\citep{hessel2018rainbow, vieillard2020munchausen, lee2025hyperspherical, palenicek2026xqc}.

Gradient TD algorithms were developed to address this divergence issue~\citep{sutton2008convergent}. This line of work led to provably convergent TD algorithms, even with nonlinear function approximation~\citep{maei2009convergent, dai2018sbeed, patterson2022generalized}. While this strong property should steer the field towards gradient TD methods, semi-gradient methods remain the preferred option.

\begin{figure}
    \centering
    \includegraphics[width=\textwidth]{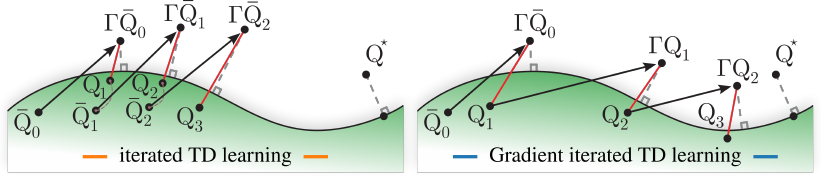}
    \vspace{-0.5cm}
    \caption{\textbf{Left}: In this figure, iterated TD learning learns $3$ projected Bellman iterations in parallel, where each Bellman iteration $\Gamma \bar{Q}_{k-1}$, built from the target estimator $\bar{Q}_{k-1}$, is learned with an online estimator $Q_k$. \textbf{Right}: Gradient iterated TD learning minimizes the sum of Bellman errors $\| \Gamma \bar{Q}_0 - Q_1 \|_2^2 + \| \Gamma Q_1 - Q_2 \|_2^2 + \| \Gamma Q_2 - Q_3 \|_2^2$. Each function $Q_k$ not only learns to regress its target $\Gamma Q_{k-1}$, but also to make the target $\Gamma Q_k$ for the following function $Q_{k+1}$ easier to regress.}
    \label{F:itd_gitd}
    \vspace{-0.3cm}
\end{figure}

Recently, iterated TD~(i-TD) learning was introduced by~\citet{vincentiterated} with the objective of improving the learning speed of TD methods by learning a sequence of action-value functions in parallel, where each function is optimized to represent the application of the Bellman operator over the previous function (see Figure~\ref{F:itd_gitd}, left). Yet, as each function tracks a moving target, the benefit of learning consecutive Bellman iterations in parallel can diminish when a function positioned at the beginning of the sequence changes faster than the following ones. This issue originates from the semi-gradient nature of this algorithm.

The objective is to develop a gradient TD method that is competitive in terms of learning speed\footnote{We are interested in achieving high returns with as few samples as possible, a property we refer to as learning speed, and treat training time as a secondary metric, as TD learning is most useful for tasks that require sample efficiency.} against semi-gradient methods by learning a sequence of action-value functions, where each function is optimized to represent the application of the Bellman operator over the previous one. To that end, we introduce \textit{Gradient Iterated Temporal-Difference}~(Gi-TD) learning, which incorporates the idea of recent gradient TD methods~\citep{ghiassian2020gradient} into the iterated TD approach~\citep{vincentiterated}. This results in an algorithm that optimizes the sequence of action-value functions as a whole, without semi-gradient updates.  

Our empirical evaluations show that Gi-TD learning: ($1$) converges on counterexamples where semi-gradient methods (including i-TD learning) fail, ($2$) performs well at scale across multiple benchmarks, demonstrating for the first time that gradient TD based methods can provide competitive learning speed in the ALE benchmark~\citep{bellemare2013arcade}, and ($3$) is particularly effective with high replay ratios, and in offline reinforcement learning.

\begin{figure}
    \centering
    \includegraphics[width=\textwidth]{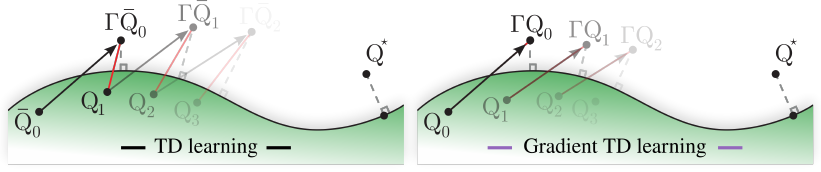}
    \vspace{-0.5cm}
    \caption{Representation of bootstrapping methods in the space of action-value functions. We represent the projection on the space of function approximation (in green) with dashed grey lines. \textbf{Left}: TD learning uses a target estimator $\bar{Q}_0$ to construct a regression target estimating the Bellman iteration $\Gamma \bar{Q}_0$, where $\Gamma$ is the Bellman operator. This quantity is learned by the online estimator $Q_1$. At every $T$ gradient steps, the target estimator is updated to the online estimator to learn the following Bellman iterations. \textbf{Right}: At every step $k$, Gradient TD learning minimizes the distance between $Q_k$ and its regression target estimating the Bellman iteration $\Gamma Q_k$, leading to a new estimate $Q_{k+1}$.}
    \label{F:td_gtd}
\end{figure}
\section{The Foundations of Temporal-Difference Learning} \label{S:foundations}
We consider a Markov Decision Process~(MDP) defined by a state space $\mathcal{S}$, an action space $\mathcal{A}$, a reward function $\mathcal{R}$, a transition kernel $\mathcal{P}$ mapping state-action pairs to probability measures on the following state, and a discount factor $\gamma$. We aim to learn the optimal action-value function $Q^*$, which represents the maximum expected sum of discounted rewards a policy can obtain by interacting with the MDP. From there, an optimal policy $\pi^*$ is obtained by choosing an action that maximizes the action-value function for each given state $s$, i.e. $\pi^*(s) \in \arg \max_a Q^*(s, a )$. Importantly, the optimal action-value function is the unique function that verifies the Bellman equation $Q = \Gamma Q$, where $\Gamma$ denotes the Bellman operator, defined as, $\Gamma Q(s, a) = \mathbb{E}_{r \sim \mathcal{R}(s, a), s' \sim \mathcal{P}(s, a)}[r + \gamma \max_{a'} Q(s', a')]$, for a state-action pair $(s, a)$. 

In problems where the reward and transition dynamics are not directly accessible, the Bellman operator must be estimated from samples. Two main bootstrapping methods can be followed to learn the optimal action-value function. Both methods rely on the contraction property of the Bellman operator, whose fixed-point is the optimal action-value function. The first and most popular direction, called TD learning, iteratively applies the Bellman operator, starting from an initial function. In an ideal setting, the Banach fixed-point theorem guarantees that this iterative process converges to the fixed point of the Bellman operator. The other direction, called Gradient TD learning, aims to minimize the distance between a $Q$-estimator $Q$ and its Bellman iteration $\Gamma Q$, thereby forcing the learned estimate to satisfy the Bellman equation.

Figure~\ref{F:td_gtd} (left) illustrates how TD learning operates in the space of action-value functions, when function approximation is used to cope with large state spaces. In the general case, the application of the Bellman operator to a $Q$-estimate $\Gamma Q$ lands outside of the space of function approximators (represented in green). TD learning starts by instantiating a $Q$-estimator $\bar{Q}_0$, commonly referred to as the target estimator, used to build a regression target $r + \gamma \max_{a'} \bar{Q}_0(s', a')$, which is a stochastic estimation of its Bellman iteration $\Gamma \bar{Q}_0$. This regression target is learned by a second estimator $Q_1$, called the online estimator, using the mean squared error. We highlight that the bar over $Q_0$ is added to emphasize that $Q_0$ is not optimized, as its parameters remain frozen. After a predefined number of gradient steps $T$, a target update is performed, which consists of setting $Q_1$ as the new target estimator by freezing its parameters, and learning a new $Q$-estimator, noted $Q_2$, which takes the role of the online estimator to minimize the distance between $\Gamma \bar{Q}_1$ and $Q_2$ \citep{mnih2015human}. 

The procedure repeats until the training stops. It benefits from the contraction property of the Bellman operator at each iteration, which brings the expected regression target closer to the optimal action-value function $Q^*$. This line of work is referred to as the semi-gradient method because the gradient with respect to the target estimator is ignored. The seminal algorithm $Q$-learning~\citep{watkins1992q} follows this paradigm, with $T$ set to $1$. While it is known that semi-gradient methods can diverge~\citep{baird1995residual}, their convergence properties are actively studied~\citep{asadi2023td}.

Figure~\ref{F:td_gtd} (right) depicts the training dynamics of Gradient TD learning. They are simpler than those of TD learning, since the objective function, i.e., the Bellman error $\| \Gamma Q - Q \|_2^2$, is optimized over all learnable parameters without stop-gradient operations. This is why convergence guarantees can be established for this approach. At each timestep $k$, the following $Q$-estimator $Q_k$ is obtained by performing a stochastic gradient descent step on the objective function, evaluated at $Q_{k-1}$. 

The difficult part of this method lies in building an unbiased estimate of the objective function's gradient. Indeed, the gradient of the Bellman error $(\Gamma Q_{\theta}(s, a) - Q_{\theta}(s, a) )^2$ is equal to $2 ~ \partial_{\theta} \Gamma Q_{\theta}(s, a) (\Gamma Q_{\theta}(s, a) - Q_{\theta}(s, a)) - 2 ~ \partial_{\theta} Q_{\theta}(s, a) (\Gamma Q_{\theta}(s, a) - Q_{\theta}(s, a))$, for a state-action pair $(s, a)$ and some parameters $\theta$. Importantly, an unbiased estimate of the second term can be obtained from a single sample $(s, a, r, s')$ by computing $- 2 ~ \partial_{\theta} Q_{\theta}(s, a) (r + \gamma \max_{a'} Q_{\theta}(s', a') - Q_{\theta}(s, a))$. Unfortunately, it is not possible to build an unbiased estimate of the first term with a single sample because it contains the product of expectations $\partial_{\theta} \Gamma Q_{\theta}(s, a) \cdot \Gamma Q_{\theta}(s, a)$. 

This is why \citet{sutton2009fast} learn the difference $\Gamma Q_{\theta} - Q_{\theta}$ with a different estimator, $H_z$, with some parameters $z$, to replace the need of two independent samples, and use the single sample that is available to estimate $\partial_{\theta} \Gamma Q_{\theta}(s, a)$. Adding a weight decay penalty on the $z$ parameters leads to the algorithm called temporal-difference learning with regularized corrections~(TDRC, \citet{ghiassian2020gradient}), where the gradients with respect to the learnable parameters $\theta$ and $z$, noted $g_\theta$ and $g_z$, are 
\begin{align}
    g_{\theta} &= \partial_{\theta} \left[ r + \gamma \max_{a'} Q_\theta(s', a') \right] H_z(s, a) - \partial_{\theta} Q_{\theta}(s, a) \left[ r + \gamma \max_{a'} Q_\theta(s', a') - Q_{\theta}(s, a) \right] \\
    g_z &= \partial_z \left[H_z(s, a) - (r + \gamma \max_{a'} Q_\theta(s', a') - Q_{\theta}(s, a)) \right]^2 + \beta ~ \partial_z \| z \|_2^2,
\end{align}
where $\beta$ is the weight decay coefficient. 

\section{Iterated Temporal-Difference Learning} \label{S:iTD}
To increase the learning speed of temporal-difference learning methods, \citet{vincentiterated} suggest to learn a sequence of $K+1$ action-value functions $(Q_k)_{k=0}^K$ where each function $Q_k$ is optimized to represent $\Gamma Q_{k-1}$, which is the application of the Bellman operator $\Gamma$ over the previous function in the sequence $Q_{k-1}$. For that, the algorithm aims at minimizing the sum of Bellman Errors~(BEs), $\sum_{k=1}^K \| \Gamma Q_{k-1} - Q_k \|_2^2$. 

The motivation for this novel objective function lies in the fact that this term is included in the upper bound on error propagation for approximate value iteration, published in~\citet{farahmand2011regularization}. Given a sequence of functions $Q_0, \hdots, Q_K$, \citet{farahmand2011regularization} derive an upper bound on the distance between the optimal action-value function $Q^*$ and the action-value function $Q^{\pi_K}$, where $\pi_K$ is the greedy policy over the last function $Q_K$ (see Theorem~\hyperref[T:error_propagation]{$3.4$} for a formal statement). This upper bound contains the weighted sum of BEs formed by the sequence of functions, i.e., $\sum_{k=1}^K \alpha_k \| \Gamma Q_{k-1} - Q_k \|^2_{2, \nu}$, where the weights $\alpha_k$ are approximated to $1$ in the objective function of i-TD learning, and $\nu$ is the distribution of the collected data. 

We present a schematic representation of i-TD learning in Figure~\ref{F:itd_gitd} (left). It considers $K$ online estimators $(Q_k)_{k=1}^K$, where each estimator $Q_k$ is represented by parameters $\theta_k$, and $K$ target estimators $(\bar{Q}_k)_{k=0}^{K-1}$, where each estimator $\bar{Q}_k$ is represented by parameters $\bar{\theta}_k$. Similarly to TD learning, each online estimator $Q_k$ learns from its respective target estimator $\bar{Q}_{k-1}$. Crucially, at every $D$ training step, a chain structure is imposed by synchronizing each target estimator $\bar{Q}_k$ with its corresponding online estimator $Q_k$, i.e., $\bar{\theta}_k \gets \theta_k$. $D$ is generally chosen significantly smaller than $T$.

Since the number of Bellman iterations learned over the course of a training is generally too large to store all $Q$-estimators in memory, $K$ is set to a reasonable value ($\approx 5$), and at every $T$ training steps, each estimator parameter $\theta_k$ is updated to the following one $\theta_{k+1}$ so that the following Bellman iterations are considered. The same is done for each target parameter $\bar{\theta}_k, k \in \{0, K-1\}$, thereby leading to learning the following Bellman iterations. This step has a similar effect to the target update in TD learning. In fact, when $K=1$, iterated TD learning reduces to TD learning.

The major limitation of i-TD learning is that each target estimator $\bar{Q}_k$ is constantly out of synchronization with its corresponding online estimator $Q_k$, which is the true moving target, as highlighted by the dashed brown curves in Figure~\ref{F:itd_gitd}. While this problem can be reduced by frequently updating each target estimator to its corresponding online estimator, it still persists even when this update is performed after each gradient step, i.e., $D=1$. This is because the target estimators are frozen during semi-gradient updates. As a result, i-TD learning minimizes the sum of BEs only under certain conditions (see Proposition~$4.1$ in \citet{vincentiterated}), which are not under the algorithm's control. i-TD learning can even lead to a diverging sum of BEs, as we show in Section~\ref{S:controlled_mdps}.

\vspace{-0.1cm}
\section{Related Work} \label{S:related_work}
\vspace{-0.1cm}
Stepping away from semi-gradient updates is not a new idea. \citet{schweitzer1985generalized} propose to minimize the Bellman error $\| \Gamma Q - Q \|$ in the context of dynamic programming. In deterministic environments, where the double sampling problem does not occur, \citet{baird1995residual} suggest scaling the gradient of the bootstrapped estimate, to be more competitive against semi-gradient methods. In the general case, the Bellman error appears less favorable as a learning objective because it is not a function of the data, which leads \citet{sutton2018reinforcement} (Section $11.6$) to conclude that this quantity is not learnable. Instead, \citet{sutton2008convergent, sutton2009fast} focus on minimizing the projected Bellman error with linear function approximators.

\citet{patterson2022generalized} generalizes this idea to nonlinear function approximation, resulting in the TDRC algorithm presented in Section~\ref{S:foundations}. We denote $\mathcal{H}$ as the class of $H$ approximators and $\mathcal{Q}$ the class of $Q$ approximators. The authors demonstrate that the algorithm's learning objective becomes closer to the projected Bellman error when $\mathcal{H}$ becomes closer to $\mathcal{Q}$, and the same objective function becomes closer to the Bellman error when $\mathcal{H}$ increases in size. For completeness, Figure~\ref{F:gtd_gitd} (left) depicts the schematic representation of TDRC with the projected Bellman error as objective function, corresponding to the case when $\mathcal{H} = \mathcal{Q}$. 

The idea of learning a sequence of action-value functions that represent consecutive Bellman iterations has been introduced by \citet{schmitt2022chaining}. \citet{vincent2024parameterized} extends the idea to nonlinear function approximation and infinite sequence length by learning a hypernetwork mapping the parameters of a $Q$-network to the parameters of the $Q$-network representing its projected Bellman iteration. Iterated TD learning (\citet{vincentiterated}, explained in Section~\ref{S:iTD}) further develops the approach to scale to deep nonlinear function approximation and learn every action-value function in parallel without requiring sequential updates. In a follow-up work, \citet{vincent2026bridging} propose a parameter-efficient version where each function in the sequence is represented by a linear head, built on top of a shared feature extractor. While showing greater learning speed on challenging benchmarks, i-TD learning suffers from moving targets, making it prone to unexpected behavior. 

Finally, the idea of learning the value improvement path~\citep{dabney2021value} is closely related to i-TD learning, differing in that it focuses on learning past Bellman iterations via auxiliary losses to improve representation learning, rather than actively learning the following Bellman iterations. 

In the following, we propose to modify i-TD learning by computing the gradient of the stochastic targets, as in TDRC, yielding Gradient Iterated TD~(Gi-TD) learning. The resulting algorithm is designed to directly minimize the sum of BEs, thereby combining the soundness of Gradient TD learning with the promise of greater learning speed coming from the idea motivating i-TD learning.

\section{Gradient Iterated Temporal-Difference Learning} \label{S:gitd}
We design an algorithm, which we call \textit{Gradient Iterated Temporal-Difference}~(Gi-TD) learning, that learns a sequence of $K+1$ action-value functions in parallel. Each function $Q_k$ is optimized to represent the Bellman iteration of the previous function in the sequence $\Gamma Q_{k-1}$. The objective function remains the same as that of i-TD learning, which is the sum of BEs $\sum_{k=1}^K \| \Gamma Q_{k-1} - Q_k \|_2^2$, but Gi-TD learning differs in the way it minimizes this sum. We note that the first function $Q_0$ is kept fixed, since it does not have a target to regress. Hence, we denote it as $\bar{Q}_0$. 

In contrast to i-TD learning, Gi-TD learning computes the gradients of the stochastic targets rather than relying on semi-gradient updates. This means that every function $Q_k$ is not only learned to approximate its target $\Gamma Q_{k-1}$, but also to make the target $\Gamma Q_k$ for the following function $Q_{k+1}$, an easier target to regress towards. It ensures that Gi-TD learning \textit{directly} minimizes the sum of BEs, as all parameters in the sum are optimized without ignoring the gradient of the targets. 

Figure~\ref{F:itd_gitd} (right) illustrates the mechanism behind the proposed algorithm for $K=3$, where the sequence of functions $\bar{Q}_0, Q_1, Q_2, Q_3$ is learned to minimize the sum of BEs $\| \Gamma \bar{Q}_0 - Q_1 \| + \| \Gamma Q_1 - Q_2 \| + \| \Gamma Q_2 - Q_3 \|$. Importantly, by directly minimizing the sum of BEs, the optimization procedure allows trade-offs between early and later Bellman errors. This means that future functions influence the learning dynamics of previous ones. This property is novel and avoids the greedy behavior of TD and i-TD learning, which only minimize the immediate Bellman error and disregard the following ones. Figures~\ref{F:td_gtd} (left) and \ref{F:itd_gitd} account for this property, representing $Q_1$ further away from $\Gamma \bar{Q}_0$ for Gi-TD learning than for TD and i-TD learning, with the following Bellman errors being smaller. 

\begin{figure}
    \centering
    \includegraphics[width=\textwidth]{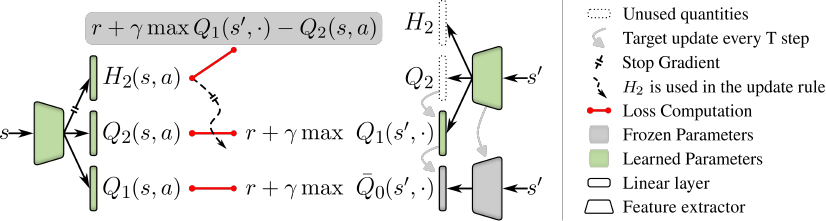}
    \caption{Training procedure of Gradient Iterated TD learning with $K=2$. $Q_1$ not only learns the regression target built from $\bar{Q}_0$, but is also optimized so that the regression target built from itself is closer to $Q_2$. $Q_2$ learns the regression target built from $Q_1$. $H_2$ approximates the difference between the regression target built from $Q_1$, and $Q_2$. To save parameters, each $Q$ and $H$-network is represented by a head, built on a shared feature extractor, reducing the number of networks to $2$.}
    \label{F:implementation_gitd}
    \vspace{-0.3cm}
\end{figure}
We now present the update rule for Gi-TD learning and derive a theoretical justification in Section~\ref{S:theoretical_results}. We represent the sequence of action-value functions with $K+1$ $Q$-estimators $\bar{Q}_0, Q_1, \hdots, Q_K$, parameterized by $\bar{\theta}_0, \theta_1, \hdots, \theta_K$. We also use $K-1$ $H$-estimators $H_2, \hdots, H_K$, parameterized by $z_2, \hdots, z_K$, where each estimator $H_k$ is trained to learn the difference $\Gamma Q_{k-1} - Q_k$. These $H$-estimators are used to compute each gradient term that includes a stochastic target $r + \gamma \max_{a'} Q_{\theta_k}(s', a')$, without requiring two independent samples, as explained in Section~\ref{S:foundations}. Following \citet{patterson2022generalized}, a weight decay is added to the parameters of the $H$-estimators. This results in the gradient estimates $\Delta_{\theta_k}$, and $\Delta_{z_k}$ to perform stochastic gradient descent on the $Q$-estimator parameters, and $H$-estimator parameters. Given a sample $(s,a,r,s')$, we define
\begin{align*}
    \Delta_{\theta_k} &= \partial_{\theta_k} \left[ r + \gamma \max_{a'} Q_{\theta_k}(s', a') \right] H_{z_{k+1}}(s, a) - \partial_{\theta_k} Q_{\theta_k}(s, a) \left[ r + \gamma \max_{a'} Q_{\theta_{k-1}}(s', a') - Q_{\theta_k}(s, a) \right] \\
    \Delta_{z_k} &= \partial_{z_k} \left[ H_{z_k}(s, a) - (r + \gamma \max_{a'} Q_{\theta_{k-1}}(s', a') - Q_{\theta_k}(s, a)) \right]^2 + \beta ~ \partial_{z_k}\|z_k\|_2^2,
\end{align*}
where $\beta$ controls the importance given to the weight decay. We note that $\Delta_{\theta_K}$ is only equal to $- \partial_{\theta_K} Q_{\theta_K}(s, a) (r + \gamma \max_{a'} Q_{\theta_{K-1}}(s', a') - Q_{\theta_K}(s, a))$, as $\theta_K$ is the last parameter of the sequence. 

Similarly to i-TD learning, a target update is performed every $T$ steps, allowing the following Bellman iterations to be considered. The implementation follows $\theta_k \gets \theta_{k+1}$, and $z_k \gets z_{k+1}$, for all possible values of $k$. An analogous update to soft-target updates~\citep{haarnoja2018soft} is also possible, where $\theta_0$ is updated to $\tau \theta_1 + (1 - \tau) \theta_0$ at every step, and $\tau$ regulates the speed at which the parameters $\theta_0$ are updated to $\theta_1$. When interacting with the environment with discrete action spaces, the action that maximizes the average prediction across the online $Q$-estimates is selected.

Figure~\ref{F:implementation_gitd} provides a summary of the training procedure with $K=2$. In practice, parameters can be shared to reduce the memory footprint, with functions represented by different heads that are built on a shared feature extractor. This reduces the number of independent estimators to $2$. We present the training procedures for TDRC and i-TD learning in Figure~\ref{F:implementation-tdrc_itd} to facilitate comparison. The pseudo-code for an instance of Gi-TD learning applied to Deep $Q$-network~(DQN, \citet{mnih2015human}) is presented in Algorithm~\ref{A:gidqn}. Algorithm~\ref{A:gisac} presents an instance of Gi-TD learning applied to Soft Actor-Critic~(SAC, \citet{haarnoja2018soft}). 

Finally, the representation capacity of the $H$-estimators can change the objective function (see Section~\ref{S:related_work}). While Figure~\ref{F:itd_gitd} (right) shows the sum of BEs as the objective function, we show in Figure~\ref{F:gtd_gitd} (right) the case where $\mathcal{H} = \mathcal{Q}$, for which the objective function is the sum of projected BEs.

\section{Analysis in Controlled MDPs} \label{S:controlled_mdps}
We start by analyzing the behavior of Gi-TD learning on the Star, Hall, and Triangle Markov Processes~(MPs)~\citep{baird1995residual, tsitsiklis1997analysis} against TD, TDRC, and i-TD learning. The Star MP, also known as the Baird's counterexample, is presented on the left of Figure~\ref{F:linear}, and the Hall MP is presented on the right. The Triangle MP is presented in Figure~\ref{F:spiral_mdp}. For each MP, the objective is to learn the value function from a dataset in which each state is visited with equal frequency. The value function is equal to zero since a reward of zero is given at every transition.

We leverage the low dimensionality of these MPs to study the behavior of the $4$ algorithms in a fully deterministic setting. This means that only one evaluation per algorithm is needed. For that, we evaluate the expected version of each algorithm, in which updates are performed synchronously across all states. We remove the influence of the $H$-network by allowing the algorithm to access the true Bellman iteration, rather than a stochastic estimate. Therefore, TDRC becomes an instantiation of the residual gradient algorithm~\citep{baird1995residual}. We also remove the influence of target updates in i-TD and Gi-TD learning and instead set $K$ to a sufficiently large value so that reward propagation is not limited. Finally, we set $T=1$ for TD learning, and $D=1$ for i-TD learning. 

For each experiment, we plot the value error of each algorithm, which is the distance between the learned estimate and the true value function, as a function of the number of gradient steps. We also report the objective function of i-TD and Gi-TD learning, which is the sum of BEs $\sum_{k=1}^K \| \Gamma V_{k-1}~-~V_k \|_2^2$, where $(V_k)_{k=0}^K$ is the considered sequence of value functions.

\vspace{-0.1cm}
\subsection{Off-policy Learning with Linear Function Approximation} \label{S:hall_mp_experiment}
\vspace{-0.1cm}
\begin{figure}
    \centering
    \includegraphics[width=\textwidth]{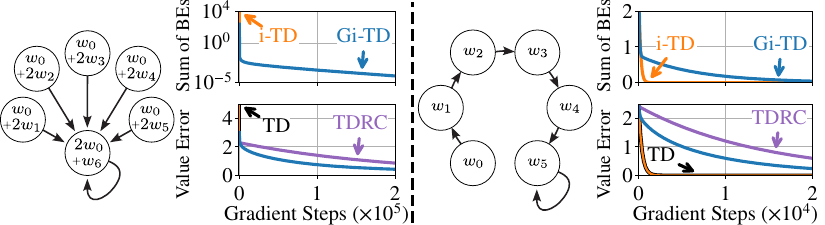}
    \caption{Evaluating the proposed approach in an \textbf{off-policy setting} with \textbf{linear function approximation}. We clarify that in the bottom plots, TD and i-TD learning overlap. \textbf{Left}: While i-TD and Gi-TD learning are both designed to decrease the sum of Bellman errors, only Gi-TD learning decreases this quantity when evaluated on Baird's counterexample. This leads to a low value error for Gi-TD learning, as opposed to i-TD learning, for which this error increases. \textbf{Right}: It is well-known that gradient TD methods have a slow learning speed when evaluated on the Hall problem~\citep{baird1995residual}. Notably, Gi-TD learning minimizes the value error faster than TDRC.}
    \label{F:linear}
\end{figure}
We first consider the Star and Hall MPs~\citep{baird1995residual} with linear function approximation (see Figure~\ref{F:linear}), in an off-policy setting. We set the learning rate to $0.08$, and $K=300$. 

For the Star MP, all weights are initialized to zero, except for $w_6$, which is set to $1$, and $\gamma = 0.99$. As expected, TD and i-TD learning diverge due to their semi-gradient nature, whereas TDRC and Gi-TD learning converge, as they are part of the Gradient TD learning paradigm (see Figure~\ref{F:linear}, left). Importantly, while the sum of BEs decreases over time for Gi-TD learning, it increases for i-TD learning, indicating that i-TD learning does not directly minimize this quantity, due to its semi-gradient nature. 

For the Hall MP, all weights are initialized to one, and $\gamma = 0.9$. \citet{baird1995residual} introduce this problem to demonstrate that semi-gradient methods can learn faster than gradient TD methods. Figure~\ref{F:linear} (right) confirms this claim. Indeed, TD and i-TD learning minimize the value error faster than TDRC and Gi-TD learning. Notably, Gi-TD learning reduces the value error faster than TDRC. Interestingly, the learning speed difference between i-TD and Gi-TD learning can be explained by the difference in the sum of BEs, supporting the relevance of this objective function.

\begin{figure}
    \centering
    \includegraphics[width=\textwidth]{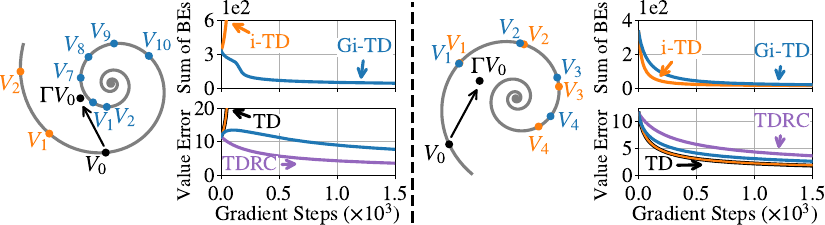}
    \caption{Evaluating the proposed approach in an \textbf{on-policy setting} with \textbf{nonlinear function approximation}, on the Triangle MP. We clarify that in the bottom plots, TD and i-TD learning overlap.  \textbf{Left}: i-TD learning increases the sum of Bellman Errors~(BEs) during training, while it is designed to minimize it. This translates to high value errors. In contrast, Gi-TD learning decreases this quantity during training, leading to a low final value error. \textbf{Right}: When changing the spiral direction, semi-gradient methods learn faster than gradient TD methods. Importantly, Gi-TD learning exhibits a faster learning speed than TDRC.}
    \label{F:non_linear}
    \vspace{-0.3cm}
\end{figure}
\subsection{On-policy Learning with Nonlinear Function Approximation}
We now focus on the Triangle MP~\citep{tsitsiklis1997analysis} in an on-policy setting, with nonlinear function approximation. We set the learning rate to $0.002$, $K=10$, $\gamma = 0.99$. All weights are initialized to $14$ so that all value functions are equal before the training starts. 

Studying this problem is helpful because the algorithm's behavior can be understood geometrically, as shown in Figures~\ref{F:td_gtd} and~\ref{F:itd_gitd}. Indeed, the Triangle MP has only $3$ states, so the value functions lie in $\mathbb{R}^3$. The space of function approximation is a spiral spanning over the plane orthogonal to the vector $(1, 1, 1)$ and including the origin. In this experiment, the Bellman operator divides the norm of any vector (which represents a value function) by $2$, and rotates the resulting vector by an angle of $-60^{\circ}$ along the axis defined by the vector $(1, 1, 1)$. This means that the application of the Bellman operator on any value function belonging to the plane mentioned earlier, remains on this plane. The true value function $(V(s_1), V(s_2), V(s_3)) = (0, 0, 0)$ is also part of this plane. Therefore, all relevant quantities lie on this plane, and the analysis can be done in $2$ dimensions. 

Figure~\ref{F:non_linear} (left) illustrates the sequence of value functions obtained at the end of training for i-TD and Gi-TD learning. We represent the initial value function $V_0$ on the plane, belonging to the space of representable value functions in gray, with the center of the spiral being the true value function $(0, 0, 0)$. As explained, $\Gamma V_0$ equals $V_0$ scaled by half and rotated by $-60^{\circ}$. In this experiment, i-TD learning tries to bring $V_1$ (in \textcolor{Orange}{orange}) closer to $\Gamma V_0$, which makes it move further away from the center of the spiral. The same goes for $V_2$, which moves further away from the center to approximate $\Gamma V_1$. This continues for all terms in the sequence of value functions, which causes the value error and the sum of BEs to increase. This experiment is another example showing that i-TD learning can lead to increasing values of its objective function. 

In contrast, Gi-TD learning exhibits a different behavior. While Gi-TD learning also brings $V_1$ (in \textcolor{RoyalBlue}{blue}) closer to $\Gamma V_0$, hence further away from the center, it makes $\Gamma V_1$ closer to $V_2$, which is initialized at the same location as $V_0$, and hence moves $V_1$ closer to the center. Overall, the sum of BEs decreases if all value functions are closer to the center of the spiral. This is why the functions move closer to the center as represented in Figure~\ref{F:non_linear} (left). This behavior translates into lower value errors and a lower sum of BEs, demonstrating the relevance of the proposed approach.

Interestingly, changing the direction of the spiral drastically alters the behavior of the studied algorithms, since the Bellman operator now points to the direction of the center following the spiral. Indeed, while i-TD learning only moves $V_1$ (in \textcolor{Orange}{orange}) closer to $\Gamma V_0$, which brings $V_1$ closer to the center, Gi-TD learning also displaces $V_1$ (in \textcolor{RoyalBlue}{blue}) such that $\Gamma V_1$ is closer to $V_2$, which is initialized at the same location as $V_0$, hence moving further away from the center. This means that semi-gradient methods have an advantage, as they ignore the gradient term coming from the bootstrapped target. Due to this additional term, the value functions learned with Gi-TD learning lag behind those learned with i-TD learning in Figure~\ref{F:non_linear} (right). Overall, all methods reduce the value error over time, with semi-gradient methods being faster than gradient TD methods. Notably, the learning speed of Gi-TD learning is higher than that of TDRC, as observed in the Hall MP experiment in Section~\ref{S:hall_mp_experiment}.

\section{Experiments} \label{S:experiments}
\vspace{-0.2cm}
We evaluate the proposed method on various benchmarks against TD, TDRC, and i-TD learning to assess its behavior in more complex environments. 

\vspace{-0.2cm}
\subsection{Setup}
\vspace{-0.1cm}
In this work, we focus on the agent's learning speed. This is why we place greater importance on the Area Under the Curve~(AUC) of the performance curve than on final performance, as it favors algorithms that continuously improve performance over those that only achieve high returns at the end of training. The AUC also has the advantage of being less dependent on the training budget. 

We use the Inter-Quantile Mean~(IQM, \citet{agarwal2021deep}) to remove outliers that may arise from lucky exploration, as the algorithms considered in this work compete at the level of optimization, given the collected data. This is done by discarding $25\%$ of each tail of the distribution of scores when aggregating results. It also allows to discard runs with faulty resets, which can occur when using the Gymnasium library~\citep{towersgymnasium}. 

When aggregating results, we report baseline-normalized scores, with the TD learning end score as baseline, rather than the human-normalized score. Indeed, TD methods learn differently than humans~\citep{delfosse2025deep}, leading to human-normalized scores that differ significantly across environments (see Figure~\ref{F:game_selection}). Therefore, aggregating human-normalized scores with the IQM metric results in discarding experiments with the lowest and highest scores, which correspond to the easiest and hardest environments. This goes against our motivation to use the IQM metric. We still use the human-normalized score for selecting a set of $10$ Atari games containing varying levels of difficulty (see Figure~\ref{F:game_selection}), as commonly done in RL~\citep{graesser2022statesparse, vincent2025eau}.

In the supplementary material, we report the list of hyperparameters in Tables~\ref{T:atari_parameters}, and~\ref{T:mujoco_parameters}, the detailed description of the aggregation protocol in Section~\ref{S:experiment_setup}, and the individual learning curves for each experiment in Section~\ref{S:individual_learning_curves}. Importantly, all shared hyperparameters are set to the same values as in the published version of the baseline for all algorithms. We set $K=5$ for i-TD and Gi-TD learning, and $\beta=1$ for TDRC and Gi-TD learning. We also set $D=1$ for i-TD learning as \citet{vincent2026bridging} show that it performs well with this value.

For experiments with discrete action spaces, we represent all functions with linear heads, built on top of a shared feature extractor for TDRC, i-TD and Gi-TD learning. For experiments with continuous action spaces, we use independent networks for all algorithms except TDRC, which uses linear heads. We ablate those choices in Section~\ref{S:ablations}. Finally, we report aggregated scores for $5$ seeds per Atari game~\citep{bellemare2013arcade} and $10$ seeds per MuJoCo environment~\citep{todorov2012}.

\begin{figure}
    \centering
    \includegraphics[width=\textwidth]{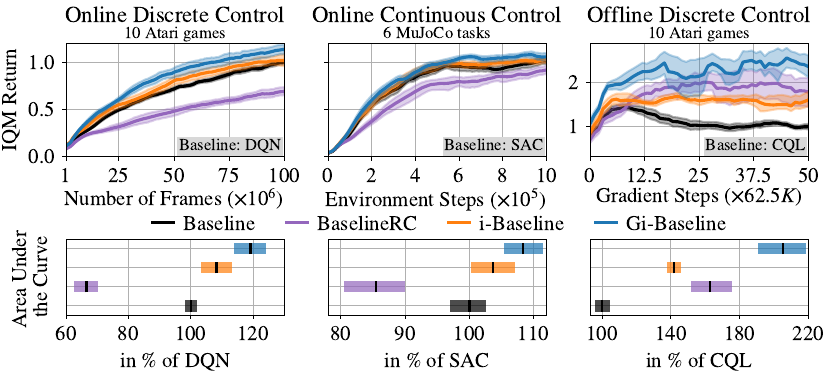}
    \caption{Evaluating the proposed approach on multiple benchmarks with \textbf{deep nonlinear function approximation}. When combined with DQN (left), SAC (center), and CQL (right), Gi-TD learning is competitive against semi-gradient methods. Importantly, the performance boost is the highest in the offline setting, where the choice of the optimization method is particularly important.}
    \label{F:performances}
\end{figure}
\subsection{Benchmark Performance} \label{S:benchmark_performance}
\textbf{Online Discrete Control.} ~We combine each of the $4$ considered TD learning methods with DQN equipped with the CNN architecture, yielding DQN, QRC, i-DQN, and Gi-DQN. We report the learning curves over $100$ million frames aggregated over $10$ Atari games in Figure~\ref{F:performances} (top, left). Remarkably, Gi-DQN outperforms all considered algorithms, showing that gradient TD methods can be competitive against semi-gradient approaches. We also report the AUC for each method, normalized by DQN's AUC, along the x-axis of Figure~\ref{F:performances} (bottom, left). The y-axis indicates which method is reported. Gi-DQN provides a $20\%$ improvement over DQN, and increases QRC's performance by $50$ points of percentage ($120 - 70 = 50$).

\textbf{Online Continuous Control.} We now investigate the ability of the $4$ considered algorithms to learn informative critics when combined with SAC. Following~\citet{haarnoja2018soft}, all networks are composed of $2$ hidden layers of $256$ neurons each. We let each algorithm interact for one million steps in $6$ MuJoCo tasks and report the aggregated SAC-normalized score in Figure~\ref{F:performances} (top, center). Here again, Gi-SAC remains competitive against the semi-gradient methods. It is not the case for SACRC, which performs worse than SAC. In Figure~\ref{F:performances} (bottom, center), the comparison of AUCs indicates that Gi-SAC achieves a $7\%$ improvement compared to SAC.

\textbf{Offline Discrete Control.} To remove the influence of exploration, we perform an evaluation in an offline scenario on $10$ Atari games. We use the dataset released by \citet{gulcehre2020rlunplugged}, which is collected by a DQN agent trained online for $200$ million interactions. This leads to a dataset of $50$ replay buffers, each comprising $4$ million frames. We combine the $4$ considered methods with Conservative $Q$-learning~(CQL, \citet{kumar2020conservative}) and let each algorithm have access to the first $10\%$ of each offline replay buffer during training. Notably, Gi-CQL outperforms the other $3$ algorithms by a large margin, as shown in Figure~\ref{F:performances} (top, right), yielding an AUC that is twice that of CQL (see Figure~\ref{F:performances}, bottom, right). Interestingly, both gradient TD methods outperform the semi-gradient methods, highlighting the benefit of using theoretically sound objective functions. 

\begin{figure}
    \centering
    \includegraphics[width=\textwidth]{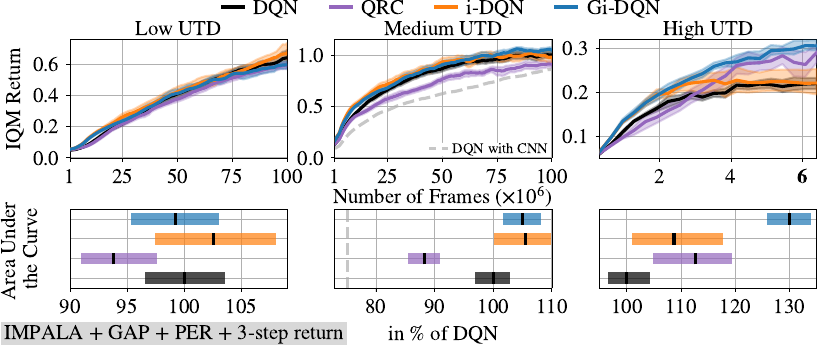}
    \caption{Evaluating the proposed approach in \textbf{combination with a strong baseline}, and with different \textbf{update-to-data~(UTD) ratios}. Gi-TD learning remains competitive against semi-gradient methods when combined with a more powerful architecture~(IMPALA + GAP), a prioritized experience replay buffer~(PER), and a $3$-step return. Notably, Gi-TD learning's learning speed is significantly greater than that of the other considered methods at high UTD ratios, a setting where the optimization method exerts a stronger influence on the learning behavior.}
    \label{F:utd_study_atari}
\end{figure}

\textbf{Scaling Up the Baseline.} We now investigate if Gi-DQN brings benefits over a stronger baseline. For that, we use the IMPALA architecture~\citep{espeholt2018impala} and add a global average pooling for each channel of the last convolutional layer, as indicated in~\citet{trumpp2025impoola, sokar2025mind}. We also implement a prioritized experience replay buffer (PER, \citet{schaul2016prioritized}) and use a $3$-step return. Figure~\ref{F:utd_study_atari} (center) confirms that the resulting baseline yields better performance than the baseline presented in Figure~\ref{F:performances} (left), as the previous baseline only reaches $75\%$ of the new baseline's AUC on the same $10$ Atari games and update-to-data ratio~(Medium $\text{UTD}=1/4$, see the grey dashed lines). In Figure~\ref{F:utd_study_atari} (center), we observe that Gi-DQN remains competitive against semi-gradient methods, improving the AUC by $5\%$ over the baseline. We stress that achieving improvements over strong baselines, such as the one we consider in this experiment, is generally harder. Similar to the previous online experiments, TDRC exhibits lower learning speed.

\textbf{Scaling Up the Replay Ratio.} Theoretically sound methods should allow for higher data utilization without the risk of divergence. Therefore, we study the behavior of the $4$ considered methods with a UTD ratio $16$ times higher than the previous one, i.e., $\text{UTD}=4$ instead of $1/4$. To reduce the computational budget, we maintain the same amount of total gradient steps, resulting in $6.25$ million interactions ($100/16 = 6.25$). In Figure~\ref{F:utd_study_atari} (right), the baseline for normalizing the learning curves is still the end performance of the strong DQN version with a medium UTD ratio to ease comparison. Remarkably, Gi-DQN outperforms the other methods, yielding an average performance $130\%$ higher than that of the strong DQN variant. In this setting, QRC also outperforms the semi-gradient methods, supporting the idea that theoretically sound methods generally perform better with more computing power. Finally, we evaluate the $4$ methods in a low UTD regime where a gradient step is performed every $32$ environment interactions, which is equal to the batch size, thereby evaluating the lowest relevant UTD ratio. As expected, all methods perform similarly, confirming that the potential of gradient TD methods lies in high UTD regimes. 

In Section~\ref{S:additional_experiments}, we perform a similar study on the MuJoCo benchmark with varying UTD values. Figure~\ref{F:utd_study_mujoco} demonstrates a similar trend, confirming the superiority of Gi-TD at high UTD ratios.

\subsection{Ablation Study} \label{S:ablations}
We ablate the design choices made in Section~\ref{S:experiments}. Specifically, we study the dependency of Gi-TD learning on the weight decay coefficient $\beta$ and the choice of the architecture. For that, we evaluate the combinations of Gi-TD learning with DQN, SAC, and CQL on $2$ Atari games with $5$ seeds, $2$ MuJoCo tasks with $5$ seeds, and $2$ Gym environments with $10$ seeds.

We run a grid search on $2$ weight decay coefficients ($\beta = 1,$ and $0.01$), and $3$ different architectures that are presented in Figure~\ref{F:architectures} (independent networks, shared feature extractor with linear heads, and with nonlinear heads), except for the experiments on Atari games, for which we do not evaluate independent architectures to reduce the computational budget. The CNN architecture is used for Gi-DQN, a $2$-layer network with $256$ neurons is used for Gi-SAC, and a $3$-layer network with $50$ neurons is used for Gi-CQL. 

To reduce dependency on other hyperparameters, we add $2$ axes to the grid and aggregate results across them. We select $2$ values of UTD ($0.25$ and $1$ for Gi-DQN, $1$ and $4$ for Gi-SAC) or $2$ dataset sizes for Gi-CQL ($10\%$ and $100\%$), and $2$ target update periods ($T=1000$ and $8000$ for Gi-DQN, $T=10$ and $1000$ for Gi-CQL) or $2$ values of soft-target updates for Gi-SAC ($\tau=0.005$ and $0.01$). 

\begin{figure}
    \centering
    \includegraphics[width=\textwidth]{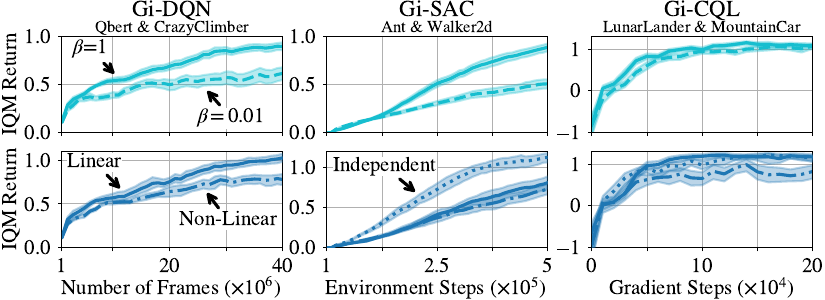}
    \caption{Ablation studies on the \textbf{weight decay coefficient} and the \textbf{architecture design choices}. \textbf{Top}: Setting $\beta=1$, as advocated in~\citet{patterson2022generalized}, yield better performance. \textbf{Bottom}: Defining a shared feature extractor with linear heads performs best for experiments with discrete actions. In the experiment with a continuous actions space (Gi-SAC), we believe that the neural networks are too shallow to benefit from parameter sharing.}
    \label{F:gi-td_ablation}
\end{figure}

Figure~\ref{F:gi-td_ablation} (top) reports the aggregated performance over architecture designs, the UTDs or dataset sizes, and target update periods or soft-target update values. It demonstrates the relevance of regularizing the $H$-networks as advocated in \citet{ghiassian2020gradient}, because setting the weight decay coefficient to a high value generally leads to better performance. Then, we report the performance of the different architectures in Figure~\ref{F:gi-td_ablation} (bottom), selecting $\beta=1$, and aggregating over the $2$ other axes. The architecture with linear heads appears to perform best, except in the continuous-action setting, where independent networks yield the best results. This explains the choices made in Section~\ref{S:experiments}. We believe that independent networks perform better for Gi-SAC because the architecture is too shallow (only $2$ layers) to benefit from parameter sharing. We also perform the same study on TDRC and i-TD learning, which we report in Figure~\ref{F:tdrc_itd_ablation}, and draw similar conclusions. 

Finally, we fix $\beta=1$, and use an architecture with linear heads to compare $2$ values of $K$ ($5$ and $50$) for i-TD and Gi-TD learning. We aggregate the learning curves across the $2$ other axes in Figure~\ref{F:K_ablation}. Notably, the dependency of Gi-TD learning on this hyperparameter is lower than for i-TD learning, which struggles to learn for $K=50$, a limitation pointed out in~\citet{vincent2026bridging}.

\section{Limitations, Conclusion \& Future Work}
Due to its more elaborate loss, the presented method requires more floating-point operations per gradient step than the other methods, which entails a longer training time (see Figure~\ref{F:params_flops}). Nonetheless, we point out that when computing the algorithm’s runtime, the environment steps are simulated, which does not reflect the real-world duration, as the simulator reduces the time required to collect data, which is generally the main bottleneck for real-world applications. 

Another limitation is that the proposed objective function is not designed to minimize the entire bound on error propagation (see Theorem~\hyperref[T:error_propagation]{$3.4$}). It only focuses on the term that depends on the sequence of action-value functions observed during training. However, the proposed method still provides some benefit compared to i-TD learning, and the other terms are related to the algorithm's exploration strategy, which is beyond the scope of this work. Finally, studying the convergence properties of the proposed approach would provide more insights into the algorithm's behavior.

To conclude, we introduced \textit{Gradient Iterated Temporal-Difference}~(Gi-TD) learning, which is designed to learn a sequence of action-value functions in parallel, where each function is optimized to represent the application of the Bellman operator over the previous function in the sequence. Its objective function is the sum of Bellman errors formed by the sequence of action-value functions. Gi-TD learning minimizes this quantity by optimizing over every learnable parameter, including those used to compute stochastic targets. Therefore, it belongs to the gradient TD line of work. Our evaluations show competitive performance against semi-gradient methods across various benchmarks, including Atari games, a result that no prior work on gradient TD methods has demonstrated.

We believe that studying a version of Gi-TD learning with different coefficients for each Bellman iteration could further boost performance. Additionally, combining Gi-TD learning with gradient eligibility traces~\citep{elelimydeep}, distributional or robust losses~\citep{bellemare2023distributional, qu2019nonlinear, patterson2022robust} is a promising step towards sample-efficient algorithms.

\clearpage

\section*{Acknowledgments}
This research was supported by “Third Wave of AI”, funded by the Excellence Program of the Hessian Ministry of Higher Education, Science, Research and Art, and by the grant “Einrichtung eines Labors des Deutschen Forschungszentrum für Künstliche Intelligenz (DFKI) an der Technischen Universität Darmstadt”. We gratefully acknowledge support from the hessian.AI Service Center (funded by the Federal Ministry of Education and Research, BMBF, grant no. 01IS22091) and the hessian.AI Innovation Lab (funded by the Hessian Ministry for Digital Strategy and Innovation, grant no. S-DIW04/0013/003).

\section*{Carbon Impact}
As recommended by \citet{lannelongue2023carbon}, we used GreenAlgorithms \citep{lannelongue2021green} and ML $CO_2$ Impact \citep{lacoste2019quantifying} to compute the carbon emission related to the production of the electricity used for the computations of our experiments. We only consider the energy used to generate the figures presented in this work and ignore the energy used for preliminary studies and for building the computing infrastructure. The estimations vary between $2.23$ and $2.29$ tonnes of $\text{CO}_2$ equivalent. As a reminder, the Intergovernmental Panel on Climate Change advocates a carbon budget of $2$ tonnes of $\text{CO}_2$ equivalent per year per person.
\bibliography{main}

\begin{thebibliography}{58}
\providecommand{\natexlab}[1]{#1}
\providecommand{\url}[1]{\texttt{#1}}
\expandafter\ifx\csname urlstyle\endcsname\relax
  \providecommand{\doi}[1]{DOI: #1}\else
  \providecommand{\doi}{DOI: \begingroup \urlstyle{rm}\Url}\fi

\bibitem[Agarwal et~al.(2021)Agarwal, Schwarzer, Castro, Courville, and Bellemare]{agarwal2021deep}
Rishabh Agarwal, Max Schwarzer, Pablo~Samuel Castro, Aaron~C Courville, and Marc Bellemare.
\newblock Deep reinforcement learning at the edge of the statistical precipice.
\newblock \emph{Advances in neural information processing systems}, 2021.

\bibitem[Asadi et~al.(2023)Asadi, Sabach, Liu, Gottesman, and Fakoor]{asadi2023td}
Kavosh Asadi, Shoham Sabach, Yao Liu, Omer Gottesman, and Rasool Fakoor.
\newblock Td convergence: An optimization perspective.
\newblock \emph{Advances in Neural Information Processing Systems}, 2023.

\bibitem[Baird(1995)]{baird1995residual}
Leemon Baird.
\newblock Residual algorithms: Reinforcement learning with function approximation.
\newblock \emph{Machine learning}, 1995.

\bibitem[Bellemare et~al.(2013)Bellemare, Naddaf, Veness, and Bowling]{bellemare2013arcade}
Marc~G Bellemare, Yavar Naddaf, Joel Veness, and Michael Bowling.
\newblock The arcade learning environment: An evaluation platform for general agents.
\newblock \emph{Journal of Artificial Intelligence Research}, 2013.

\bibitem[Bellemare et~al.(2023)Bellemare, Dabney, and Rowland]{bellemare2023distributional}
Marc~G Bellemare, Will Dabney, and Mark Rowland.
\newblock \emph{Distributional reinforcement learning}.
\newblock MIT Press, 2023.

\bibitem[Bradbury et~al.(2018)Bradbury, Frostig, Hawkins, Johnson, Leary, Maclaurin, Necula, Paszke, Vander{P}las, Wanderman-{M}ilne, and Zhang]{jax2018github}
James Bradbury, Roy Frostig, Peter Hawkins, Matthew~James Johnson, Chris Leary, Dougal Maclaurin, George Necula, Adam Paszke, Jake Vander{P}las, Skye Wanderman-{M}ilne, and Qiao Zhang.
\newblock \emph{{JAX}: composable transformations of {P}ython+{N}um{P}y programs}, 2018.

\bibitem[Brockman et~al.(2016)Brockman, Cheung, Pettersson, Schneider, Schulman, Tang, and Zaremba]{brockman2016openai}
Greg Brockman, Vicki Cheung, Ludwig Pettersson, Jonas Schneider, John Schulman, Jie Tang, and Wojciech Zaremba.
\newblock Openai gym.
\newblock \emph{arXiv preprint arXiv:1606.01540}, 2016.

\bibitem[Castro et~al.(2018)Castro, Moitra, Gelada, Kumar, and Bellemare]{castro2018dopamine}
Pablo~Samuel Castro, Subhodeep Moitra, Carles Gelada, Saurabh Kumar, and Marc~G Bellemare.
\newblock Dopamine: A research framework for deep reinforcement learning.
\newblock \emph{arXiv preprint arXiv:1812.06110}, 2018.

\bibitem[Dabney et~al.(2021)Dabney, Barreto, Rowland, Dadashi, Quan, Bellemare, and Silver]{dabney2021value}
Will Dabney, Andr{\'e} Barreto, Mark Rowland, Robert Dadashi, John Quan, Marc~G Bellemare, and David Silver.
\newblock The value-improvement path: Towards better representations for reinforcement learning.
\newblock In \emph{Proceedings of the AAAI conference on artificial intelligence}, 2021.

\bibitem[Dai et~al.(2018)Dai, Shaw, Li, Xiao, He, Liu, Chen, and Song]{dai2018sbeed}
Bo~Dai, Albert Shaw, Lihong Li, Lin Xiao, Niao He, Zhen Liu, Jianshu Chen, and Le~Song.
\newblock Sbeed: Convergent reinforcement learning with nonlinear function approximation.
\newblock In \emph{International Conference on Machine Learning}, 2018.

\bibitem[Degrave et~al.(2022)Degrave, Felici, Buchli, Neunert, Tracey, Carpanese, Ewalds, Hafner, Abdolmaleki, de~Las~Casas, et~al.]{degrave2022magnetic}
Jonas Degrave, Federico Felici, Jonas Buchli, Michael Neunert, Brendan Tracey, Francesco Carpanese, Timo Ewalds, Roland Hafner, Abbas Abdolmaleki, Diego de~Las~Casas, et~al.
\newblock Magnetic control of tokamak plasmas through deep reinforcement learning.
\newblock \emph{Nature}, 2022.

\bibitem[Delfosse et~al.(2025)Delfosse, Bl{\"u}ml, Tatai, Vincent, Gregori, Dillies, Peters, Rothkopf, and Kersting]{delfosse2025deep}
Quentin Delfosse, Jannis Bl{\"u}ml, Fabian Tatai, Th{\'e}o Vincent, Bjarne Gregori, Elisabeth Dillies, Jan Peters, Constantin~A Rothkopf, and Kristian Kersting.
\newblock Deep reinforcement learning agents are not even close to human intelligence.
\newblock In \emph{Inductive Biases in Reinforcement Learning Workshop @RLC}, 2025.

\bibitem[Elelimy et~al.(2025)Elelimy, Daley, Patterson, Machado, White, and White]{elelimydeep}
Esraa Elelimy, Brett Daley, Andrew Patterson, Marlos~C Machado, Adam White, and Martha White.
\newblock Deep reinforcement learning with gradient eligibility traces.
\newblock In \emph{Reinforcement Learning Journal}, 2025.

\bibitem[Elsayed et~al.(2024)Elsayed, Vasan, and Mahmood]{elsayed2024streaming}
Mohamed Elsayed, Gautham Vasan, and A~Rupam Mahmood.
\newblock Streaming deep reinforcement learning finally works.
\newblock \emph{arXiv preprint arXiv:2410.14606}, 2024.

\bibitem[Espeholt et~al.(2018)Espeholt, Soyer, Munos, Simonyan, Mnih, Ward, Doron, Firoiu, Harley, Dunning, et~al.]{espeholt2018impala}
Lasse Espeholt, Hubert Soyer, Remi Munos, Karen Simonyan, Vlad Mnih, Tom Ward, Yotam Doron, Vlad Firoiu, Tim Harley, Iain Dunning, et~al.
\newblock Impala: Scalable distributed deep-rl with importance weighted actor-learner architectures.
\newblock In \emph{International Conference on Machine Learning}, 2018.

\bibitem[Farahmand(2011)]{farahmand2011regularization}
Amir-massoud Farahmand.
\newblock Regularization in reinforcement learning.
\newblock \emph{University of Alberta}, 2011.

\bibitem[Ghiassian et~al.(2020)Ghiassian, Patterson, Garg, Gupta, White, and White]{ghiassian2020gradient}
Sina Ghiassian, Andrew Patterson, Shivam Garg, Dhawal Gupta, Adam White, and Martha White.
\newblock Gradient temporal-difference learning with regularized corrections.
\newblock In \emph{International Conference on Machine Learning}, 2020.

\bibitem[Graesser et~al.(2022)Graesser, Evci, Elsen, and Castro]{graesser2022statesparse}
Laura Graesser, Utku Evci, Erich Elsen, and Pablo~Samuel Castro.
\newblock The state of sparse training in deep reinforcement learning.
\newblock In \emph{International Conference on Machine Learning}, 2022.

\bibitem[Gulcehre et~al.(2020)Gulcehre, Wang, Novikov, Paine, G{\'o}mez, Zolna, Agarwal, Merel, Mankowitz, Paduraru, et~al.]{gulcehre2020rlunplugged}
Caglar Gulcehre, Ziyu Wang, Alexander Novikov, Thomas Paine, Sergio G{\'o}mez, Konrad Zolna, Rishabh Agarwal, Josh~S Merel, Daniel~J Mankowitz, Cosmin Paduraru, et~al.
\newblock Rl unplugged: A suite of benchmarks for offline reinforcement learning.
\newblock \emph{Advances in Neural Information Processing Systems}, 2020.

\bibitem[Haarnoja et~al.(2018)Haarnoja, Zhou, Abbeel, and Levine]{haarnoja2018soft}
Tuomas Haarnoja, Aurick Zhou, Pieter Abbeel, and Sergey Levine.
\newblock Soft actor-critic: Off-policy maximum entropy deep reinforcement learning with a stochastic actor.
\newblock In \emph{International Conference on Machine Learning}, 2018.

\bibitem[Harris et~al.(2020)Harris, Millman, van~der Walt, Gommers, Virtanen, Cournapeau, Wieser, Taylor, Berg, Smith, Kern, Picus, Hoyer, van Kerkwijk, Brett, Haldane, del R{\'{i}}o, Wiebe, Peterson, G{\'{e}}rard-Marchant, Sheppard, Reddy, Weckesser, Abbasi, Gohlke, and Oliphant]{harris2020array}
Charles~R. Harris, K.~Jarrod Millman, St{\'{e}}fan~J. van~der Walt, Ralf Gommers, Pauli Virtanen, David Cournapeau, Eric Wieser, Julian Taylor, Sebastian Berg, Nathaniel~J. Smith, Robert Kern, Matti Picus, Stephan Hoyer, Marten~H. van Kerkwijk, Matthew Brett, Allan Haldane, Jaime~Fern{\'{a}}ndez del R{\'{i}}o, Mark Wiebe, Pearu Peterson, Pierre G{\'{e}}rard-Marchant, Kevin Sheppard, Tyler Reddy, Warren Weckesser, Hameer Abbasi, Christoph Gohlke, and Travis~E. Oliphant.
\newblock Array programming with {NumPy}.
\newblock \emph{Nature}, 2020.

\bibitem[Hessel et~al.(2018)Hessel, Modayil, Van~Hasselt, Schaul, Ostrovski, Dabney, Horgan, Piot, Azar, and Silver]{hessel2018rainbow}
Matteo Hessel, Joseph Modayil, Hado Van~Hasselt, Tom Schaul, Georg Ostrovski, Will Dabney, Dan Horgan, Bilal Piot, Mohammad Azar, and David Silver.
\newblock Rainbow: Combining improvements in deep reinforcement learning.
\newblock In \emph{Proceedings of the AAAI conference on artificial intelligence}, 2018.

\bibitem[Jiang et~al.(2022)Jiang, Zhang, Chelu, White, and van Hasselt]{jiang2022learning}
Ray Jiang, Shangtong Zhang, Veronica Chelu, Adam White, and Hado van Hasselt.
\newblock Learning expected emphatic traces for deep rl.
\newblock In \emph{Proceedings of the AAAI conference on artificial intelligence}, 2022.

\bibitem[Kingma \& Ba(2015)Kingma and Ba]{kingma2015adam}
Diederik Kingma and Jimmy Ba.
\newblock Adam: A method for stochastic optimization.
\newblock In \emph{International Conference on Learning Representations}, 2015.

\bibitem[Kumar et~al.(2020)Kumar, Zhou, Tucker, and Levine]{kumar2020conservative}
Aviral Kumar, Aurick Zhou, George Tucker, and Sergey Levine.
\newblock Conservative q-learning for offline reinforcement learning.
\newblock In \emph{Advances in Neural Information Processing Systems}, 2020.

\bibitem[Lacoste et~al.(2019)Lacoste, Luccioni, Schmidt, and Dandres]{lacoste2019quantifying}
Alexandre Lacoste, Alexandra Luccioni, Victor Schmidt, and Thomas Dandres.
\newblock Quantifying the carbon emissions of machine learning.
\newblock \emph{arXiv preprint arXiv:1910.09700}, 2019.

\bibitem[Lannelongue \& Inouye(2023)Lannelongue and Inouye]{lannelongue2023carbon}
Lo{\"\i}c Lannelongue and Michael Inouye.
\newblock Carbon footprint estimation for computational research.
\newblock \emph{Nature Reviews Methods Primers}, 2023.

\bibitem[Lannelongue et~al.(2021)Lannelongue, Grealey, and Inouye]{lannelongue2021green}
Lo{\"\i}c Lannelongue, Jason Grealey, and Michael Inouye.
\newblock Green algorithms: quantifying the carbon footprint of computation.
\newblock \emph{Advanced Science}, 2021.

\bibitem[Lee et~al.(2025)Lee, Lee, Seno, Kim, Stone, and Choo]{lee2025hyperspherical}
Hojoon Lee, Youngdo Lee, Takuma Seno, Donghu Kim, Peter Stone, and Jaegul Choo.
\newblock Hyperspherical normalization for scalable deep reinforcement learning.
\newblock In \emph{International Conference on Machine Learning}, 2025.

\bibitem[Machado et~al.(2018)Machado, Bellemare, Talvitie, Veness, Hausknecht, and Bowling]{machado2018revisiting}
Marlos~C Machado, Marc~G Bellemare, Erik Talvitie, Joel Veness, Matthew Hausknecht, and Michael Bowling.
\newblock Revisiting the arcade learning environment: Evaluation protocols and open problems for general agents.
\newblock \emph{Journal of Artificial Intelligence Research}, 2018.

\bibitem[Maei et~al.(2009)Maei, Szepesvari, Bhatnagar, Precup, Silver, and Sutton]{maei2009convergent}
Hamid Maei, Csaba Szepesvari, Shalabh Bhatnagar, Doina Precup, David Silver, and Richard~S Sutton.
\newblock Convergent temporal-difference learning with arbitrary smooth function approximation.
\newblock \emph{Advances in neural information processing systems}, 2009.

\bibitem[Mnih et~al.(2015)Mnih, Kavukcuoglu, Silver, Rusu, Veness, Bellemare, Graves, Riedmiller, Fidjeland, Ostrovski, et~al.]{mnih2015human}
Volodymyr Mnih, Koray Kavukcuoglu, David Silver, Andrei~A Rusu, Joel Veness, Marc~G Bellemare, Alex Graves, Martin Riedmiller, Andreas~K Fidjeland, Georg Ostrovski, et~al.
\newblock Human-level control through deep reinforcement learning.
\newblock \emph{nature}, 2015.

\bibitem[Palenicek et~al.(2026)Palenicek, Vogt, Watson, Posner, and Peters]{palenicek2026xqc}
Daniel Palenicek, Florian Vogt, Joe Watson, Ingmar Posner, and Jan Peters.
\newblock Xqc: Well-conditioned optimization accelerates deep reinforcement learning.
\newblock \emph{International Conference on Learning Representations}, 2026.

\bibitem[Patterson et~al.(2022{\natexlab{a}})Patterson, Liao, and White]{patterson2022robust}
Andrew Patterson, Victor Liao, and Martha White.
\newblock Robust losses for learning value functions.
\newblock \emph{IEEE Transactions on Pattern Analysis and Machine Intelligence}, 2022{\natexlab{a}}.

\bibitem[Patterson et~al.(2022{\natexlab{b}})Patterson, White, and White]{patterson2022generalized}
Andrew Patterson, Adam White, and Martha White.
\newblock A generalized projected bellman error for off-policy value estimation in reinforcement learning.
\newblock \emph{Journal of Machine Learning Research}, 2022{\natexlab{b}}.

\bibitem[Qu et~al.(2019)Qu, Mannor, and Xu]{qu2019nonlinear}
Chao Qu, Shie Mannor, and Huan Xu.
\newblock Nonlinear distributional gradient temporal-difference learning.
\newblock In \emph{International Conference on Machine Learning}, 2019.

\bibitem[Schaul et~al.(2016)Schaul, Quan, Antonoglou, and Silver]{schaul2016prioritized}
Tom Schaul, John Quan, Ioannis Antonoglou, and David Silver.
\newblock Prioritized experience replay.
\newblock In \emph{International Conference on Learning Representations}, 2016.

\bibitem[Schmitt et~al.(2022)Schmitt, Shawe-Taylor, and Van~Hasselt]{schmitt2022chaining}
Simon Schmitt, John Shawe-Taylor, and Hado Van~Hasselt.
\newblock Chaining value functions for off-policy learning.
\newblock In \emph{Proceedings of the AAAI Conference on Artificial Intelligence}, 2022.

\bibitem[Schrittwieser et~al.(2020)Schrittwieser, Antonoglou, Hubert, Simonyan, Sifre, Schmitt, Guez, Lockhart, Hassabis, Graepel, et~al.]{schrittwieser2020mastering}
Julian Schrittwieser, Ioannis Antonoglou, Thomas Hubert, Karen Simonyan, Laurent Sifre, Simon Schmitt, Arthur Guez, Edward Lockhart, Demis Hassabis, Thore Graepel, et~al.
\newblock Mastering atari, go, chess and shogi by planning with a learned model.
\newblock \emph{Nature}, 2020.

\bibitem[Schweitzer \& Seidmann(1985)Schweitzer and Seidmann]{schweitzer1985generalized}
Paul~J Schweitzer and Abraham Seidmann.
\newblock Generalized polynomial approximations in markovian decision processes.
\newblock \emph{Journal of Mathematical Analysis and Applications}, 1985.

\bibitem[Sokar \& Castro(2025)Sokar and Castro]{sokar2025mind}
Ghada Sokar and Pablo~Samuel Castro.
\newblock Mind the gap! the challenges of scale in pixel-based deep reinforcement learning.
\newblock \emph{Advances in Neural Information Processing Systems}, 2025.

\bibitem[Sutton \& Barto(2018)Sutton and Barto]{sutton2018reinforcement}
Richard Sutton and Andrew Barto.
\newblock \emph{Reinforcement learning: an introduction}.
\newblock MIT press Cambridge, 2018.

\bibitem[Sutton(1988)]{sutton1988learning}
Richard~S Sutton.
\newblock Learning to predict by the methods of temporal differences.
\newblock \emph{Machine learning}, 1988.

\bibitem[Sutton et~al.(2008)Sutton, Szepesv{\'a}ri, and Maei]{sutton2008convergent}
Richard~S Sutton, Csaba Szepesv{\'a}ri, and Hamid~Reza Maei.
\newblock A convergent o (n) algorithm for off-policy temporal-difference learning with linear function approximation.
\newblock \emph{Advances in neural information processing systems}, 2008.

\bibitem[Sutton et~al.(2009)Sutton, Maei, Precup, Bhatnagar, Silver, Szepesv{\'a}ri, and Wiewiora]{sutton2009fast}
Richard~S Sutton, Hamid~Reza Maei, Doina Precup, Shalabh Bhatnagar, David Silver, Csaba Szepesv{\'a}ri, and Eric Wiewiora.
\newblock Fast gradient-descent methods for temporal-difference learning with linear function approximation.
\newblock In \emph{International Conference on Machine Learning}, 2009.

\bibitem[Todorov et~al.(2012)Todorov, Erez, and Tassa]{todorov2012}
Emanuel Todorov, Tom Erez, and Yuval Tassa.
\newblock Mujoco: A physics engine for model-based control.
\newblock In \emph{International Conference on Intelligent Robots and Systems}, 2012.

\bibitem[Towers et~al.(2026)Towers, Kwiatkowski, Balis, De~Cola, Deleu, Goul{\~a}o, Andreas, Krimmel, KG, Perez-Vicente, et~al.]{towersgymnasium}
Mark Towers, Ariel Kwiatkowski, John~U Balis, Gianluca De~Cola, Tristan Deleu, Manuel Goul{\~a}o, Kallinteris Andreas, Markus Krimmel, Arjun KG, Rodrigo De~Lazcano Perez-Vicente, et~al.
\newblock Gymnasium: A standard interface for reinforcement learning environments.
\newblock In \emph{Advances in Neural Information Processing Systems Datasets and Benchmarks Track}, 2026.

\bibitem[Trumpp et~al.(2025)Trumpp, Sch{\"a}fftlein, Theile, and Caccamo]{trumpp2025impoola}
Raphael Trumpp, Ansgar Sch{\"a}fftlein, Mirco Theile, and Marco Caccamo.
\newblock Impoola: The power of average pooling for image-based deep reinforcement learning.
\newblock \emph{Reinforcement Learning Journal}, 2025.

\bibitem[Tsitsiklis \& Van~Roy(1997)Tsitsiklis and Van~Roy]{tsitsiklis1997analysis}
John~N Tsitsiklis and Benjamin Van~Roy.
\newblock An analysis of temporal-difference learning with function approximation.
\newblock \emph{IEEE Transactions on Automatic Control}, 1997.

\bibitem[Vasan et~al.(2024)Vasan, Elsayed, Azimi, He, Shariar, Bellinger, White, and Mahmood]{vasan2024avg}
Gautham Vasan, Mohamed Elsayed, Alireza Azimi, Jiamin He, Fahim Shariar, Colin Bellinger, Martha White, and A.~Rupam Mahmood.
\newblock Deep policy gradient methods without batch updates, target networks, or replay buffers.
\newblock In \emph{Advances in Neural Information Processing Systems}, 2024.

\bibitem[Vieillard et~al.(2020)Vieillard, Pietquin, and Geist]{vieillard2020munchausen}
Nino Vieillard, Olivier Pietquin, and Matthieu Geist.
\newblock Munchausen reinforcement learning.
\newblock \emph{Advances in Neural Information Processing Systems}, 2020.

\bibitem[Vincent et~al.(2024)Vincent, Metelli, Belousov, Peters, Restelli, and D'Eramo]{vincent2024parameterized}
Th{\'e}o Vincent, Alberto~Maria Metelli, Boris Belousov, Jan Peters, Marcello Restelli, and Carlo D'Eramo.
\newblock Parameterized projected bellman operator.
\newblock In \emph{Proceedings of the AAAI Conference on Artificial Intelligence}, 2024.

\bibitem[Vincent et~al.(2025{\natexlab{a}})Vincent, Faust, Tripathi, Peters, and D'Eramo]{vincent2025eau}
Th{\'e}o Vincent, Tim Faust, Yogesh Tripathi, Jan Peters, and Carlo D'Eramo.
\newblock Eau de $ q $-network: Adaptive distillation of neural networks in deep reinforcement learning.
\newblock \emph{Reinforcement Learning Journal}, 2025{\natexlab{a}}.

\bibitem[Vincent et~al.(2025{\natexlab{b}})Vincent, Palenicek, Belousov, Peters, and D'Eramo]{vincentiterated}
Th{\'e}o Vincent, Daniel Palenicek, Boris Belousov, Jan Peters, and Carlo D'Eramo.
\newblock Iterated $ q $-network: Beyond one-step bellman updates in deep reinforcement learning.
\newblock \emph{Transactions on Machine Learning Research}, 2025{\natexlab{b}}.

\bibitem[Vincent et~al.(2026)Vincent, Tripathi, Faust, Oren, Peters, and D'Eramo]{vincent2026bridging}
Th{\'e}o Vincent, Yogesh Tripathi, Tim Faust, Yaniv Oren, Jan Peters, and Carlo D'Eramo.
\newblock Bridging the performance gap between target-free and target-based reinforcement learning.
\newblock \emph{International Conference on Learning Representations}, 2026.

\bibitem[Watkins \& Dayan(1992)Watkins and Dayan]{watkins1992q}
Christopher~JCH Watkins and Peter Dayan.
\newblock Q-learning.
\newblock \emph{Machine learning}, 1992.

\bibitem[Wurman et~al.(2022)Wurman, Barrett, Kawamoto, MacGlashan, Subramanian, Walsh, Capobianco, Devlic, Eckert, Fuchs, et~al.]{wurman2022outracing}
Peter~R Wurman, Samuel Barrett, Kenta Kawamoto, James MacGlashan, Kaushik Subramanian, Thomas~J Walsh, Roberto Capobianco, Alisa Devlic, Franziska Eckert, Florian Fuchs, et~al.
\newblock Outracing champion gran turismo drivers with deep reinforcement learning.
\newblock \emph{Nature}, 2022.

\bibitem[Young \& Tian(2019)Young and Tian]{young19minatar}
Kenny Young and Tian Tian.
\newblock Minatar: An atari-inspired testbed for thorough and reproducible reinforcement learning experiments.
\newblock \emph{arXiv preprint arXiv:1903.03176}, 2019.

\end{thebibliography}
\bibliographystyle{rlj}

\beginSupplementaryMaterials
\appendix

\section*{Table of Contents} \label{S:appendix_TOC}
\begin{list}{}{\leftmargin=1.5em \labelwidth \leftmargin \labelsep=0.5em \itemsep=0em}
    \vspace{-0.5cm}
    \item \contentsline{section}{\numberline {A} \hyperref[S:theoretical_results]{\color{black} Theoretical Results}}{\pageref{S:theoretical_results}}{}
    \vspace{-0.4cm}
    \item \contentsline{section}{\numberline {B} \hyperref[S:schematic_representations]{\color{black} Schematic Representations}}{\pageref{S:schematic_representations}}{}
    \vspace{-0.4cm}
    \item \contentsline{section}{\numberline {C} \hyperref[S:algorithmic_details]{\color{black} Algorithmic Details}}{\pageref{S:algorithmic_details}}{}
    \vspace{-0.4cm}
    \item \contentsline{section}{\numberline {D} \hyperref[S:experiment_setup]{\color{black} Experiment Setup}}{\pageref{S:experiment_setup}}{}
    \vspace{-0.4cm}
    \item \contentsline{section}{\numberline {E} \hyperref[S:additional_experiments]{\color{black} Additional Experiments}}{\pageref{S:additional_experiments}}{}
    \vspace{-0.5cm}
    \begin{list}{}{\leftmargin=2.5em \labelwidth \leftmargin \labelsep=0.5em \itemsep=0em}
        \item \contentsline{subsection}{\numberline {E.1} \hyperref[S:scaling_up_utd_sac]{\color{black}  Scaling Up the Replay Ratio in Continuous Actions Domains}}{\pageref{S:scaling_up_utd_sac}}{}
        \item \contentsline{subsection}{\numberline {E.2} \hyperref[S:additional_ablation_studies]{\color{black} Additional Ablation Studies}}{\pageref{S:additional_ablation_studies}}{}
    \end{list}
    \vspace{-0.8cm}
    \item \contentsline{section}{\numberline {F} \hyperref[S:individual_learning_curves]{\color{black} Individual Learning Curves}}{\pageref{S:individual_learning_curves}}{}
\end{list}

\section{Theoretical Results} \label{S:theoretical_results}
\begin{theoremfarahmand*} \label{T:error_propagation}
Let $N \in \mathbb{N}^*$, and $\rho$, $\nu$ two probability measures on $\mathcal{S} \times \mathcal{A}$. For any sequence $(Q_k)_{k=0}^K \in \mathcal{Q}_{\Theta}^{K+1}$ where $R_{\gamma}$ depends on the reward function and the discount factor, we have
\begin{equation*}
    \| Q^* - Q^{\pi_K} \|_{1, \rho} \leq C_{K, \gamma, R_{\gamma}} + \inf_{r \in [0, 1]} F(r; K, \rho, \nu, \gamma) \bigg( \textstyle\sum_{k=1}^{K} \alpha_k^{2r} \| \Gamma^*Q_{k - 1} - Q_k \|_{2, \nu}^{2} \bigg)^{\frac{1}{2}}
\end{equation*}
where $C_{K, \gamma, R_{\gamma}}$, $F(r; K, \rho, \nu, \gamma)$, and $(\alpha_k)_{k=1}^K$ do not depend on $(Q_k)_{k=1}^K$. $\pi_K$ is a greedy policy computed from $Q_K$.
\end{theoremfarahmand*}

\textbf{Formal Derivation of Gi-TD learning's update rule.} ~We present a theoretical justification for Gi-TD learning's update rule to complement the intuitive explanation given in Section~\ref{S:gitd}. The goal is to reformulate the objective function of Gi-TD learning such that the double sampling problem is replaced by a min-max problem, as in \citet{patterson2022generalized}. Similar to \citet{sutton2009fast}, Gi-TD learning's update rule can then be derived using a saddle-point update and gradient correction.

The objective function of Gi-TD learning is the sum of Bellman Errors~(BEs) formed by the sequence of considered action-value functions $(Q_{\theta_k})_{k=0}^K$. For each $k$, and state-action-reward-next-state $(s, a, r, s')$, we note $\delta_k(s, a, r, s') \coloneq r + \gamma \max_{a'}Q_{\theta_{k-1}}(s', a') - Q_{\theta_k}(s, a)$. Therefore, Gi-TD learning's objective is to minimize the sum of BEs that can be written as
\begin{equation*}
    \sum_{k=1}^K \sum_{(s, a) \in \mathcal{S} \times \mathcal{A}} d(s, a) ~ \mathbb{E}_{r\sim \mathcal{R}(s, a), s' \sim \mathcal{P}(s, a)} \left[ \delta_k(s, a, r, s') \right]^2,
\end{equation*}
where $d(s, a)$ is the state-action distribution.

Unfortunately, estimating this quantity with a single sample yields a biased gradient, since the square of expectations requires two independent samples to be estimated without bias. This is why we reformulate the objective function using the fact that the square function is equal to its biconjugate, i.e., $x^2 = \max_{h \in \mathbb{R}} 2xh - h^2, \forall x \in \mathbb{R}$. 

Starting from the original formulation, we replace the square function by its biconjugate to obtain the saddle-point formulation. We note $\delta_k^{\mathbb{E}}(s, a)  \coloneq \mathbb{E}_{r\sim \mathcal{R}(s, a), s' \sim \mathcal{P}(s, a)}\left[ \delta_k(s, a, r, s') \right]$, and write:
\begin{align*}
    \sum_{k=1}^K \sum_{(s, a) \in \mathcal{S} \times \mathcal{A}} d(s, a) \delta_k^{\mathbb{E}}(s, a)^2
    &= \sum_{k=1}^K \sum_{(s, a) \in \mathcal{S} \times \mathcal{A}} d(s, a) \max_{h \in \mathbb{R}} 2 ~ h ~ \delta_k^{\mathbb{E}}(s, a) - h^2 \\    
    &\overset{\mathrm{(a)}}{=} \sum_{k=1}^K \max_{H \in \mathcal{F}_{\text{all}}} \sum_{(s, a) \in \mathcal{S} \times \mathcal{A}} d(s, a) \left( 2 ~ H(s, a) ~ \delta_k^{\mathbb{E}}(s, a) - H(s, a)^2 \right) \\ 
    &\overset{\mathrm{(b)}}{=} \max_{H \in \mathcal{F}^K_{\text{all}}} \sum_{k=1}^K \sum_{(s, a) \in \mathcal{S} \times \mathcal{A}} d(s, a) \left( 2 ~ H_k(s, a) ~ \delta_k^{\mathbb{E}}(s, a) - H_k(s, a)^2 \right),
\end{align*}
where $\mathcal{F}_{\text{all}}$ represents the space of all functions mapping state-action pairs to real values. We remark that Equations $(a)$ and $(b)$ are true because of the interchangeability of the summation and the max in this situation. Indeed, Equation $(a)$ holds because a different value of $H(s, a)$ can be chosen for each state-action pair $(s, a)$. Similarly, Equation $(b)$ holds because a different value of $H_k(s, a)$ can be chosen for each state-action pair $(s, a)$, and for each $k$.

We remark that the sum of BEs can be converted to the sum of projected BEs when $\mathcal{F}_{\text{all}} = \mathcal{Q}$, where $\mathcal{Q}$ is the space of function approximators, as explained in \citet{patterson2022generalized}.

From there, different update rules can be established to perform stochastic gradient descent on the min-max problem. We use $K$ functions $(H_{z_k})_{k=1}^K$, parameterized by $(z_k)_{k=1}^K$, to represent the quantity optimized by the maximum operator. Overall, the update rule for all learnable parameters is
\begin{align*}
    \theta_k &\gets \theta_k - \eta_{\theta} ~ \Delta_{\theta_k}, \text{ for } k \in \{0, \hdots, K\} \\
    z_k &\gets z_k - \eta_{z} ~ \Delta_{z_k}, \text{ for } k \in \{1, \hdots, K\}, \\
\end{align*}
where $\eta_{\theta}$, and $\eta_{z}$ are the learning rates shared across all values for $k$. In the following, we derive the gradient estimates $\Delta_{\theta_k}$ and $\Delta_{z_k}$. 

Following GTD2~\citep{sutton2009fast}, the gradient estimates for saddle-point updates are
\begin{align*}
    \Delta_{\theta_k} 
    &= \partial_{\theta_k} \left[ \sum_{j=1}^K 2 ~ H_{z_j}(s, a) ~ \delta_j(s, a, r, s') - H_{z_j}(s, a)^2 \right], \text{ where } \delta_j \text{ is a function of } \theta_{j - 1} \text{ and } \theta_j,\\
    &= 
    \begin{cases}
        2 ~ H_{z_1}(s, a) ~ \partial_{\theta_0} \delta_{1}(s, a, r, s')
        &\text{ if } k = 0, \\
        2 ~ H_{z_k}(s, a) ~ \partial_{\theta_k} \delta_k(s, a, r, s') + 2 ~ H_{z_{k+1}}(s, a) ~ \partial_{\theta_k} \delta_{k+1}(s, a, r, s')
        &\text{ if } k \in \{1, \hdots, K-1\}, \\
        2 ~ H_{z_K}(s, a) ~ \partial_{\theta_k} \delta_K(s, a, r, s')
        &\text{ if } k = K, \\
    \end{cases} \\
    &=
    \begin{cases}
        2 ~ H_{z_1}(s, a) ~ \partial_{\theta_0} \left( r + \gamma \max_{a'}Q_{\theta_0}(s', a') \right)
        &\text{ if } k = 0, \\
        -2 ~ H_{z_k}(s, a)~ \partial_{\theta_k} Q_{\theta_k}(s, a) + 2 ~ H_{z_{k+1}}(s, a)~ \partial_{\theta_k} \left( r + \gamma \max_{a'}Q_{\theta_k}(s', a') \right)
        &\text{ if } k \in \{1, \hdots, K-1\}, \\
        -2 ~ H_{z_K}(s, a) ~ \partial_{\theta_K} Q_{\theta_K}(s, a) 
        &\text{ if } k = K, \\
    \end{cases} \\
    \Delta_{z_k} 
    &= -\partial_{z_k} \left[ \sum_{j=1}^K \left( 2 ~ H_{z_j}(s, a) ~ \delta_j(s, a, r, s') - H_{z_j}(s, a)^2 \right) \right] \\ 
    &= - \left[ 2 ~ \delta_k(s, a, r, s') ~ \partial_{z_k} H_{z_k}(s, a) - 2 ~ H_{z_k}(s, a) ~ \partial_{z_k}H_{z_k}(s, a) \right] \\
    &= 2 ~ \left( H_{z_k}(s, a) - \delta_k(s, a, r, s') \right) \partial_{z_k} H_{z_k}(s, a) \\
    &= \partial_{z_k} \left[ H_{z_k}(s, a) - \delta_k(s, a, r, s') \right]^2 \\
    &= \partial_{z_k} \left[ H_{z_k}(s, a) - (r + \gamma \max_{a'}Q_{\theta_{k-1}}(s', a') - Q_{\theta_k}(s, a)) \right]^2.
\end{align*}

Finally, we reduce the influence of the learned estimates $(H_{z_k})_{k=1}^K$ on the update rule of the action-value function parameters. For that, following the TDC algorithm~\citep{sutton2009fast}, we incorporate a gradient correction term in $\Delta_{\theta_k}$. Additionally, we use a weight decay penalty regulated with the coefficient $\beta$ in $\Delta_{z_k}$, as advocated in \citet{ghiassian2020gradient}. 

We also freeze the first parameters $\theta_0$. Indeed, $K$ can only be set to a small value due to memory constraints. This value is generally far lower than the number of Bellman iterations required to ensure sufficient reward propagation. This means that target updates have to be implemented by setting $\theta_k \gets \theta_{k+1}$, and $z_k \gets z_{k+1}$ every $T$ steps. By freezing the first action-value function in the sequence, we ensure it stays in place at the position that has been learned, and is not influenced by the following functions in the sequence that are still being optimized.

When incorporating gradient correction, weight decay penalty, and a frozen $\theta_0$, we obtain the gradient estimates presented in Section~\ref{S:gitd}:
\begin{align*}
    \Delta_{\theta_k} 
    &= 
    \begin{cases}
        \begin{aligned}
            &- ( r + \gamma \max_{a'}Q_{\theta_{k-1}}(s', a') - Q_{\theta_k}(s, a) ) ~ \partial_{\theta_k} Q_{\theta_k}(s, a) \\
            &\quad + H_{z_{k+1}}(s, a) ~ \partial_{\theta_k} ( r + \gamma \max_{a'}Q_{\theta_k}(s', a') )
        \end{aligned}
        &\text{if } k \in \{1, \hdots, K-1\}, \\
        - ( r + \gamma \max_{a'}Q_{\theta_{k-1}}(s', a') - Q_{\theta_k}(s, a) ) ~ \partial_{\theta_K} Q_{\theta_K}(s, a)
        &\text{if } k = K,
    \end{cases} \\
    \Delta_{z_k} 
    &= \partial_{z_k} \left( H_{z_k}(s, a) - (r + \gamma \max_{a'}Q_{\theta_{k-1}}(s', a') - Q_{\theta_k}(s, a)) \right)^2 + \beta ~ \partial_{z_k} \| z_k \|_2^2,
\end{align*}
where a factor of $2$ is ignored in all terms of $\Delta_{\theta_k}$ for clarity.

Finally, we remark that $H_{z_1}$ does not appear thanks to the gradient correction and the fact that $\theta_0$ is frozen. Therefore, only $K-1$ $H$-parameters are needed to compute the update rule.

\section{Schematic Representations} \label{S:schematic_representations}
\begin{figure}[H]
    \centering
    \includegraphics[width=\textwidth]{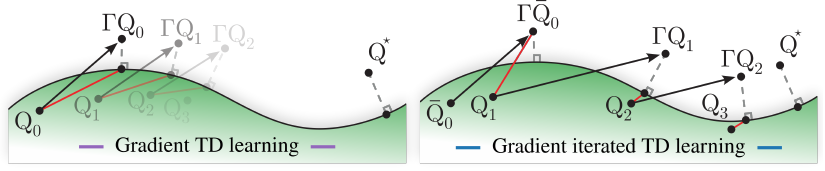}
    \caption{\textbf{Left}: As opposed to Figures~\ref{F:td_gtd} (right), which shows the case where Gradient TD learning minimizes the Bellman error, this figure depicts the case where Gradient TD learning minimizes the projected Bellman error. This case happens when the classes of function approximators for $H$-networks and $Q$-networks are equal. \textbf{Right}: Similarly, in this figure, gradient iterated TD learning minimizes the sum of projected Bellman errors, and not the sum of Bellman errors as shown in Figure~\ref{F:itd_gitd} (right). Importantly, the first Bellman error is not projected as $\bar{Q}_0$ is kept frozen.}
    \label{F:gtd_gitd}
\end{figure}

\begin{figure}[H]
    \centering
    \includegraphics[width=\textwidth]{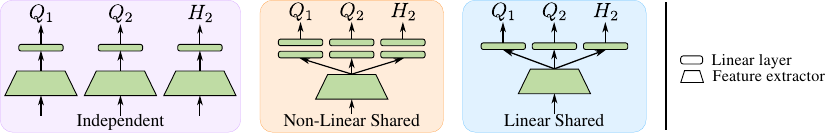}
    \caption{Different architecture choices. \textbf{Left}: Each action-value function or functions $H$ uses a separate network. \textbf{Center}: All action-value functions and functions $H$ use a shared feature extractor with heads that are composed of two linear layers with a ReLU function in between. \textbf{Right}: All action-value functions and functions $H$ use a shared feature extractor with linear heads.}
    \label{F:architectures}
\end{figure}

\clearpage
\begin{figure}[H]
    \centering
    \includegraphics[width=\textwidth]{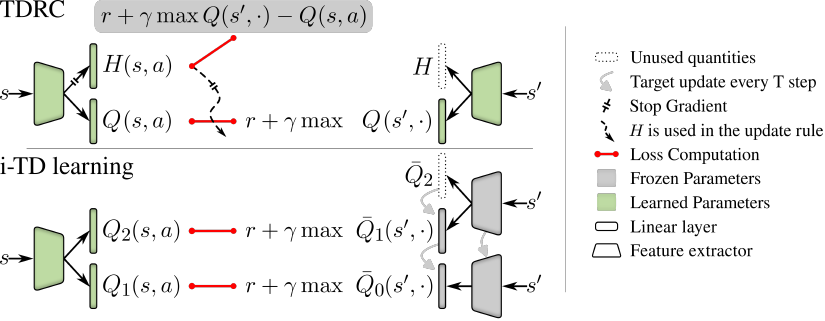}
    \caption{\textbf{Top}: Training procedure of \textbf{TDRC} with a shared architecture, and $2$ heads. The $2$ heads represent the action-value function $Q$, and the $H$-prediction $H$. The $Q$-head regresses a target built from itself. The $H$-head regresses the difference between the regression target and the $Q$-head. \textbf{Bottom}: Training procedure for \textbf{i-TD learning} for $K=2$, with shared architecture. Each $Q_k$ regresses a target built from $Q_{k-1}$. For building regression targets, parameters are kept frozen, depicted in grey on the right-hand side.}
    \label{F:implementation-tdrc_itd}
\end{figure}

\section{Algorithmic Details} \label{S:algorithmic_details}
Our codebase relies on the JAX framework~\citep{jax2018github}, the Adam optimizer~\citep{kingma2015adam}, and the NumPy library~\citep{harris2020array}. It is available at:
\begin{itemize}
    \setlength{\itemindent}{0.9cm}
    \item[\raisebox{-0.1cm}{\includegraphics[height=0.9\baselineskip]{figures/github_logo.png}}] \url{https://github.com/theovincent/Gi-DQN}, for the experiments with DQN.
    \item[\raisebox{-0.1cm}{\includegraphics[height=0.9\baselineskip]{figures/github_logo.png}}] \url{https://github.com/theovincent/Gi-SAC}, for the experiments with SAC.
    \item[\raisebox{-0.1cm}{\includegraphics[height=0.9\baselineskip]{figures/github_logo.png}}] \url{https://github.com/theovincent/Gi-CQL}, for the experiments with CQL.
\end{itemize}
\begin{figure}[H]
    \centering
    \includegraphics[width=\textwidth]{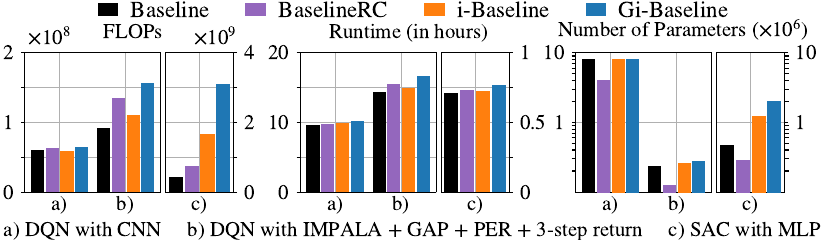}
    \caption{Comparison of computational requirements between TD, TDRC, i-TD, and Gi-TD learning. The comparison is made for the $2$ different architectures used for the online Atari experiment, and for the Multi-Layer Perceptron~(MLP) used for the MuJoCo experiment. All algorithms use a shared feature extractor on top of linear heads, except for i-SAC and Gi-SAC, which use independent networks. We do not report the metrics for the offline experiments as they are similar to those of the online experiments. \textbf{Left}: Number of Floating-Points Operations~(FLOPs) required for each gradient step. As expected, gradient TD methods require more FLOPs as they compute the gradient of the target. \textbf{Center}: Training time of the different methods computed on an NVIDIA GeForce RTX $4090$ with the game Breakout for the DQN experiments, and with the task Ant for the SAC experiments. Importantly, while a difference is noticeable in terms of FLOPs, the training times are similar. Gi-TD only requires a few minutes more than the other methods. \textbf{Right}: When sharing parameters, Gi-TD uses a similar amount of parameters compared to the other methods.}
    \label{F:params_flops}
\end{figure}

\clearpage

\begin{algorithm}[H]
\caption{\textit{Gradient iterated} Deep $Q$-Network (Gi-DQN). Modifications to DQN are in \textcolor{RoyalBlue}{blue}.}
\label{A:gidqn}
\begin{algorithmic}[1]
\State Initialize \textcolor{RoyalBlue}{$K$} $Q$-network parameters $\textcolor{RoyalBlue}{(}\theta_k\textcolor{RoyalBlue}{)_{k = 1}^K}$, \textcolor{RoyalBlue}{$K-1$ $H$-network parameters $(z_k)_{k = 2}^K$}, and an empty replay buffer $\mathcal{D}$. Initialize the frozen target $Q$-network parameters $\theta_0 \gets \theta_1$.
\State \textbf{Repeat}
\Indent
\State Take action $a_t \sim \epsilon\text{-greedy}(\textcolor{RoyalBlue}{\frac{1}{K}\sum_{k=1}^K} Q_{\theta_{k}})$; Observe reward $r_t$, next state $s_{t+1}$.
\State Update $\mathcal{D} \leftarrow \mathcal{D} \bigcup \{(s_t, a_t, r_t, s_{t+1})\}$.
\State \textbf{for} UTD \textbf{updates}
\Indent
\State Sample a mini-batch $\mathcal{B} = \{ (s, a, r, s') \}$ from $\mathcal{D}$.
\State Set $\delta_{k}(d) = r + \gamma \max_{a'} Q_{\theta_{k-1}}(s', a') - Q_{\theta_k}(s, a)$, for $d$ $\in$ $\mathcal{B}$, and \textcolor{RoyalBlue}{$k \in \left\{1,\dots,K\right\}$}
\State Compute the losses, for $d \in \mathcal{B}$: \Comment{$\lceil \cdot \rceil$ indicates a stop gradient operation.}

    $\quad\begin{aligned}
    \mathcal{L}_Q(d) &= \textcolor{RoyalBlue}{\sum_{k=1}^{K-1} \left[ (r+\gamma \max_{a'} Q_{\theta_k}(s',a')) \cdot \lceil H_{z_{{k+1}}}(s,a) \rceil - Q_{\theta_{k}}(s,a) \cdot  \lceil \delta_k(d) \rceil  \right]} \\
    &\quad - Q_{\theta_K}(s,a) \cdot \lceil \delta_K(d) \rceil
    \end{aligned}$

    $\quad\begin{aligned}
    \textcolor{RoyalBlue}{\mathcal{L}_H(d) = \sum_{k=2}^{K} \left[ \lceil \delta_k(d) \rceil - H_{z_k}(s,a)\right] ^2 + \beta ||z_k||_2^2}.
    \end{aligned}$
    
\State Update $\theta_k$ with $\nabla_{\theta_k} \sum_{d \in \mathcal{B}}\mathcal{L}_Q(d)$, \textcolor{RoyalBlue}{for $k$ $\in \{1,...K-1\}$.} 
\State \textcolor{RoyalBlue}{Update $z_k$ with $\nabla_{z_k}$$\sum_{d \in \mathcal{B}}\mathcal{L}_H(d)$, for $k$ $\in \{2, ..., K-1\}$.}
\EndIndent
\State \textbf{every $T$ steps}
\Indent
\State $\theta_{k} \gets \theta_{k+1}$, \textcolor{RoyalBlue}{for $ k \in \{0,\dots, K-1\}$.}
\State \textcolor{RoyalBlue}{$z_k \gets z_{k+1}$, for $k \in \{2,\dots, K-1\}$.}
\EndIndent
\EndIndent
\end{algorithmic}
\end{algorithm}
\vspace{-1cm}
\begin{algorithm}[H]
\caption{\textit{Gradient iterated} Soft Actor Critic (Gi-SAC). Modifications to SAC are in \textcolor{RoyalBlue}{blue}.}
\label{A:gisac}
\begin{algorithmic}[1]
\State Initialize the policy network parameters $\phi$, $2$ $\times$ \textcolor{RoyalBlue}{$K$} critic-network parameters $\textcolor{RoyalBlue}{(}\theta_k^1,\theta_k^2\textcolor{RoyalBlue}{)_{k = 1}^K}$, \textcolor{RoyalBlue}{$2 \times$ $(K-1)$ $H$-network parameters $(z_k^1, z_k^2)_{k = 2}^K$}, an empty replay buffer $\mathcal{D}$, and $\alpha$ the entropy coefficient. Initialize the $2$ frozen target critic-networks parameters $\theta^i_0 \leftarrow \theta^i_1$ for $i=1, 2$.
\State \textbf{Repeat}
\Indent
\State Take action $a_t \sim \pi_\phi(\cdot|s_t)$; Observe reward $r_t$, next state $s_{t+1}$.
\State Update $\mathcal{D} \gets \mathcal{D} \bigcup \{(s_t, a_t, r_t, s_{t+1})\}$.
\State \textbf{for} UTD \textbf{updates}
\Indent
\State Sample a mini-batch $\mathcal{B} = \{ (s, a, r, s') \}$ from $\mathcal{D}$.
\State $\begin{aligned}[t]
&\delta^i_{k}(d) =  r + \gamma \min_{j \in \{1, 2\}} Q_{\theta_{k-1}^j}(s', a') - \gamma \alpha \log \pi_\phi(a' | s') - Q_{\theta^i_k}(s, a)\text{,}\\
&\text{where }  a' \sim \pi_\phi(\cdot|s_t),\text{for } d \in \mathcal{B}, \textcolor{RoyalBlue}{k \in \left\{1,\dots,K\right\}}\text{, and } i \in \{1, 2\}.
\end{aligned}$
\State Compute the critic losses, for $d \in \mathcal{B}$, and $i \in \{1, 2\}$: \Comment{$\lceil \cdot \rceil$ indicates stop gradient.}

    $\quad\begin{aligned}
        \mathcal{L}_{Q}^i(d) &= \textcolor{RoyalBlue}{\sum_{k=1}^{K-1} \left[ \left(r + \gamma \min_{j \in \{1, 2\}} Q_{\theta_{k-1}^j}(s', a') 
         - \gamma \alpha \log \pi_\phi(a' | s') \right) \cdot \lceil H_{z^i_{{k+1}}}(s,a) \rceil \right. } \\ 
        &\quad \textcolor{RoyalBlue}{\left. - Q_{\theta^i_{k}}(s,a) \cdot  \lceil \delta^i_k(d) \rceil \right] } \textcolor{black}{- Q_{\theta^i_K}(s,a) \cdot \lceil \delta^i_K(d) \rceil}\text{, where } a' \sim \pi_\phi(\cdot|s_t)
    \end{aligned}$

    $\quad \begin{aligned}
    \textcolor{RoyalBlue}{\mathcal{L}_{H}^i(d) = \sum_{k=2}^{K} \left[ \lceil \delta^i_k(d) \rceil - H_{z^i_k}(s,a) \right] ^2 + \beta ||z^i_k||_2^2}.
    \end{aligned}$

\State Compute the actor loss, for $d \in \mathcal{B}$: 
    
    $\quad \begin{aligned}
    \mathcal{L}_{\pi}(d) &= \min_{i \in \{1, 2\}} \textcolor{RoyalBlue}{\frac{1}{K} \sum_{k=1}^{K}}Q_{\theta_{k}^i}(s, a) - \alpha \log \pi_\phi(a | s), \\
    &\quad \text{where } a \sim \pi_\phi(\cdot | s), \textcolor{RoyalBlue}{\text{for }k \in \{1,...K-1\}}, \text{ and } i \in \{1, 2\}.
    \end{aligned}$

\State Update $\theta^i_k$ with $\nabla_{\theta_k} \sum_{d \in \mathcal{B}}\mathcal{L}^i_{Q}(d)$, \textcolor{RoyalBlue}{for $k$ $\in \{1,...K-1\}$}, and $i \in \{1, 2\}$.
\State \textcolor{RoyalBlue}{Update $z^i_k$ with $\nabla_{z_k}$$\sum_{d \in \mathcal{B}}\mathcal{L}^i_{H}(d)$, for $k$ $\in \{1,...K-1\}$, and $i \in \{1, 2\}$.}
\State Update $\phi$ from $\nabla_\phi \sum_{d \in \mathcal{B}}\mathcal{L}_{\pi}(d)$.
\State Update $\theta_0^i \gets \tau \theta_1^i + (1 - \tau) \theta_0^i$, for $i \in \{1, 2\}$.
\EndIndent

\EndIndent
\end{algorithmic}
\end{algorithm}

\begin{figure}[H]
    \begin{lstlisting}[language=Python]      
    import jax
    import jax.numpy as jnp


    (*@\textcolor{orange}{def}@*) (*@\textcolor{blue}{loss\_on\_single\_sample}@*)(params, target_params, sample):
        # Compute (Q_1, ..., Q_K), (H_2, ..., H_K) for state s,
        # with shape (K, n_actions), (K-1, n_actions).
        q_s, h_s = network.apply(params, sample.state)
        
        # Compute Q_0 for next state s', with shape (1, n_actions).
        q_first_next_s = target_network.apply(target_params, sample.next_state)
        
        # Compute (Q_1, ..., Q_K) for next state s', with shape (K, n_actions).
        q_following_next_s, _ = network.apply(params, sample.next_state)

        q_next_s = jnp.concatenate(q_first_next_s, q_following_next_s[:-1])
        targets = sample.reward + gamma * q_next_s.max((*@\textcolor{red}{axis}@*)=1)
        td_errors = targets - q_s[:, sample.action]

        loss_targets = targets[1:] * jax.lax.stop_gradient(h_s)
        loss_online = -q_s[:, sample.action] * jax.lax.stop_gradient(td_errors)
        loss_h = jnp.square(h_s[:, sample.action] - jax.lax.stop_gradient(td_errors[1:]))

        (*@\textcolor{orange}{return}@*) loss_targets.sum() + loss_online.sum() + loss_h.sum()
        \end{lstlisting}
    \caption{\textbf{Loss computation in JAX} of Gi-DQN for a single sample. Importantly, the loss for the $Q$-networks and $H$-networks can be computed in a single function using vector operations. The variable \texttt{network} is a neural network instantiated to output the values for all the action-value functions except the first one, $\bar{Q}_0$, and the $H$-values. The weights parameterizing this network are called \texttt{params}. The variable \texttt{target\_network} outputs the values representing the first action-value function $\bar{Q}_0$. It uses the weights \texttt{target\_params}.}
    \label{fig:gidqn_python_code}
\end{figure}

\section{Experiment Setup} \label{S:experiment_setup}
\begin{wrapfigure}{r}{0.25\textwidth}
    \vspace{-0.4cm}
    \centering
    \includegraphics[width=0.194909091\textwidth]{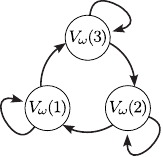}
    \caption{Triangle Markov Process, consisting of three states. The value each state $i$ is approximated by $V_{\omega}(i)$, for some parameters $\omega$.}
    \label{F:spiral_mdp}
    \vspace{-1.2cm}
\end{wrapfigure}
\textbf{Triangle MP Setup.} ~The Triangle MP is presented in Figure~\ref{F:spiral_mdp}. It is composed of $3$ states. The space of function approximation is a spiral lying on a plane in the space of value function. For a given parameter $\omega$, the associated value function is defined by
\begin{equation*}
    \begin{pmatrix}
           V_{\omega}(1) \\
           V_{\omega}(2) \\
           V_{\omega}(3)
         \end{pmatrix} 
    = e^{0.15 ~\omega} \cdot \left[ 
    \cos(0.866 ~\omega) 
    \begin{pmatrix}
        1 \\
        0 \\
        -1
    \end{pmatrix}
    - \sin(0.866 ~\omega) \frac{\epsilon}{\sqrt{3}}
    \begin{pmatrix}
        1 \\
        -2 \\
        1
    \end{pmatrix}
    \right],
\end{equation*}
where $\epsilon$ determines the direction of rotation. $\epsilon$ is equal to $-1$ in Figure~\ref{F:non_linear} (left), and $1$ in Figure~\ref{F:non_linear} (right).

\textbf{Atari Setup.} ~We build our codebase following \citet{machado2018revisiting} standards and taking inspiration from \citet{castro2018dopamine} and \citet{vincentiterated} codebases\footnote{\url{https://github.com/google/dopamine}, and \url{https://github.com/slimRL/slimDQN}}. Namely, we use the \textit{game over} signal to terminate an episode instead of the life signal. The input given to the neural network is a concatenation of $4$ frames in grayscale of dimension $84$ by $84$. To get a new frame, we sample $4$ frames from the Gym environment \citep{brockman2016openai} configured with no frame-skip, and apply a max pooling operation on the $2$ last grayscale frames. We use sticky actions to make the environment stochastic (with $p = 0.25$).

\begin{wrapfigure}{l}{0.5\textwidth}
    \vspace{0.15cm}
    \centering
    \includegraphics[width=0.4\textwidth]{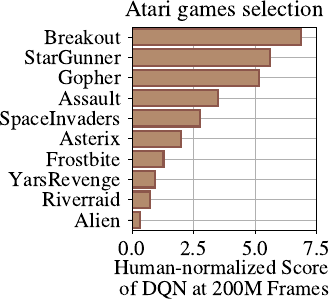}
    \caption{Human-normalized score of $10$ Atari games obtained by a DQN agent after collecting for $200$ million frames.}
    \label{F:game_selection}
    \vspace{-0.2cm}
\end{wrapfigure}

\textbf{Atari Games Selection.} ~Our evaluations are performed on $10$ games, which are chosen to represent a large variety of human-normalized scores,  reached by DQN agent trained on $200$ million frames, as shown in Figure~\ref{F:game_selection}. The human-normalized scores are computed from human and random scores that are reported in \citet{schrittwieser2020mastering}. For the offline experiment, we used the datasets provided by \citet{gulcehre2020rlunplugged}. The codebase for the offline experiments is adapted from the code released by~\citet{vincent2025eau}\footnote{\url{https://github.com/slimRL/slimCQL}}.

\textbf{MuJoCo Setup.} ~Our codebase is a fork from the codebase released by \citet{vincent2025eau}\footnote{\url{https://github.com/slimRL/slimSAC}}. It relies on the Gym library~\citep{towersgymnasium}.

\textbf{Performance Aggregation.} ~In most figures in this work, we aggregate results over seeds and environments to account for the stochasticity of the training process and the variation in the training dynamics over the different environments. For that, we first extract the undiscounted returns collected during training for each seed. Then, we smooth the signal for readability by replacing each value with the average over a window of size $5$ that covers the neighboring collected returns. We note the resulting metric for an algorithm $a$ at timestep $t$, for a seed $i$, and an environment $j$, $s_{i, j}^a(t)$. 

We stress that the collected returns are not those obtained during a separate evaluation phase, as focusing on the returns obtained during training is closer to the initial motivation behind online learning~\citep{machado2018revisiting}.

We leverage the Inter-Quantile Mean~\citep{agarwal2021deep} to aggregate the collected returns. Before computing the IQM over all runs, we normalize each collected return $s_{i, j}^a(t)$ by the average end performance of the baseline in the same environment. Therefore, the learning curve report for each timestep $t$, IQM$\left(\left\{\frac{s_{i, j}^a(t)}{\frac{1}{\text{n\_seeds}}\sum_is_{i, j}^b(L)}, \text{for all $i$ and $j$} \right\} \right)$, where $L$ is the last timestep of the training. The computation of the $95\%$ stratified bootstrap confidence intervals follows the same methodology.

We also report the normalized IQM Area Under the Curve~(AUC). For that, we compute the normalized scores as described before, sum them across all timesteps, and compute the IQM. We then divide the obtained values by the baseline IQM AUC to facilitate comparison.

\begin{table}[H]
\noindent
\begin{minipage}[t]{0.49\textwidth}
    \centering
    \caption{Summary of hyperparameters used for the Atari experiments. We note $\text{Conv}_{a,b}^d C$ a 2D convolutional layer with $C$ filters of size $a \times b$ and of stride $d$, and $\text{FC }E$ a fully connected layer with $E$ neurons.}\label{T:atari_parameters}
    \begin{tabular}{ l | r }
        \toprule
        \multicolumn{2}{ c }{Environment} \\ 
        \hline
        Discount factor $\gamma$ & $0.99$ \\
        \hline
        Horizon & $27\,000$ \\
        \hline
        Full action space & No \\
        \hline 
        Reward clipping & clip($-1, 1$) \\
        \midrule
        \multicolumn{2}{ c }{Shared hyperparameters} \\       
        \hline    
        Batch size & $32$ \\
        \hline    
        \multirow{3}{*}{Torso architecture CNN} & $\text{Conv}_{8,8}^4 32$ \\
        & $- \text{Conv}_{4,4}^2 64$ \\
        & $- \text{Conv}_{3,3}^1 64$ \\
        \hline    
        \multirow{3}{*}{Torso architecture IMPALA} & $\text{Conv}_{8,8}^4 16$ \\
        & $- \text{Conv}_{4,4}^2 32$ \\
        & $- \text{Conv}_{3,3}^1 32$ \\
        \hline
        \multirow{2}{*}{Head architecture} & $ \text{FC }512$\\
        & $- \text{FC }n_{\mathcal{A}}$ \\
        \hline   
        Activations & ReLU \\
        \midrule
        \multicolumn{2}{ c }{Online Training} \\
        \hline
        Target update & \multirow{2}{*}{$8\,000$} \\
        period $T$ & \\
        \hline 
        Initial number & \multirow{2}{*}{$20\,000$} \\
        of samples & \\
        \hline 
        Maximum replay & \multirow{2}{*}{$1\,000\,000$} \\
        buffer capacity & \\
        \hline
        Number of training & \multirow{2}{*}{$250\,000$} \\
        steps per epoch & \\
        \hline    
        Starting $\epsilon$ & $1$ \\
        \hline    
        Ending $\epsilon$ & $10^{-2}$ \\
        \hline
        $\epsilon$ linear decay & \multirow{2}{*}{$250\,000$} \\
        duration & \\
        \hline    
        Learning rate & $6.25 \times 10^{-5}$ \\
        \hline    
        Adam $\epsilon$ & $1.5 \times 10^{-4}$ \\
        \bottomrule
        
        \multicolumn{2}{ c }{Offline Training} \\
        \hline
        Target update & \multirow{2}{*}{$2\,000$} \\
        period $T$ & \\
        \hline
        CQL $\alpha$ & 0.1 \\
        \hline    
        Learning rate & $5 \times 10^{-5}$ \\
        \hline    
        Adam $\epsilon$ & $3.125 \times 10^{-4}$ \\
        \hline
        Number of training & \multirow{2}{*}{$62\,500$} \\
        steps per epoch & \\
        \hline
        Replay buffer size & $5\,000\,000$ \\
        \midrule      
    \end{tabular}
\end{minipage}
\hfill
\begin{minipage}[t]{0.49\textwidth}
    \centering
    \caption{Summary of all hyperparameters used for the MuJoCo experiments. We note $\text{FC }E$ a fully connected layer with $E$ neurons.} \label{T:mujoco_parameters}
    \begin{tabular}{ l | r }
        \toprule
        \multicolumn{2}{ c }{Environment} \\ 
        \hline
        Discount factor $\gamma$ & $0.99$ \\
        \hline
        Horizon & $1\,000$ \\
        \hline
        \multicolumn{2}{ c }{Shared Hyperparameters} \\ 
        \hline 
        Initial number & \multirow{2}{*}{$5\,000$} \\
        of samples & \\
        \hline 
        Maximum replay & \multirow{2}{*}{$1\,000\,000$} \\
        buffer capacity & \\
        \hline   
        Batch size & $256$ \\
        \hline    
        Learning rate & $10^{-3}$ \\
        \hline 
        Policy delay & $1$ \\
        \hline 
        Actor and critic & $\text{FC }256$ \\
        architecture & $- \text{FC }256$ \\
        \hline 
        Soft-target update $\tau$  & $5 \times 10^{-3}$ \\
        \hline    
        Adam $\epsilon$ & $10^{-8}$ \\
        \bottomrule
    \end{tabular}
\end{minipage}
\end{table}

\begin{table}[H]
    \caption{Summary of hyperparameters used for the \textbf{offline} ablations on LunarLander \textbf{(LL)} and MountainCar \textbf{(MC)}. We note $\text{FC }E$ a fully connected layer with $E$ neurons.}\label{T:LL_MC_parameters}
\begin{minipage}[t]{0.49\textwidth}
    \centering
    \begin{tabular}{ l | r }
        \toprule
        \multicolumn{2}{ c }{Environments} \\ 
        \hline
        Discount factor $\gamma$ & $0.99$ \\
        \hline
        \multirow{2}{*}{Horizon} & $500$ (LL) \\
         &$200$ (MC) \\
        \midrule
        \multicolumn{2}{ c }{Offline Training} \\
        \hline
        CQL $\alpha$ & 0.1 \\
        \hline    
        Learning rate & $5 \times 10^{-4}$ \\
        \hline    
        Adam $\epsilon$ & $1 \times 10^{-8}$ \\
        \hline
        Number of training & \multirow{2}{*}{$10\,000$} \\
        steps per epoch & \\
        \hline    
        Batch size & $32$ \\
        \hline    
        \multirow{2}{*}{Torso architecture} & FC $50$ \\
        &-FC $50$\\
        \hline
        \multirow{2}{*}{Head architecture} & -FC $50 $ \\
        & $ \text{-FC }n_{\mathcal{A}}$ \\ 
        \midrule  
    \end{tabular}
\end{minipage}
\hfill
\begin{minipage}[t]{0.49\textwidth}
    \centering
    \begin{tabular}{ l | r }
    \multicolumn{2}{ c }{Sample collection with DQN} \\ 
        \hline
        Number of initial & \multirow{2}{*}{$1\,000$} \\
        samples \\
        \hline 
        Number of training & \multirow{2}{*}{$10\,000$}\\
        steps per epoch \\
        \hline
        Maximum replay & \multirow{2}{*}{$10\,000$} \\
        buffer capacity \\
        \hline
        Target update  & $100$ (LL) \\
        period & $200$ (MC) \\ 
        \hline    
        Batch Size & $64$ (LL),  $32$ (MC) \\
        \hline 
        UTD & $1$ \\
        \hline    
        Starting $\epsilon$ & $1$ \\
        \hline    
        Ending $\epsilon$ & $10^{-2}$ \\
        \hline
        \multirow{2}{*}{Features} & FC $200$ (LL)\\
        &FC $50$ (MC) \\
        \hline
        \multirow{2}{*}{Torso architecture} & -FC $200$ (LL) \\
        &-FC $50$ (MC) \\
        & $ \text{-FC }n_{\mathcal{A}}$ \\ 
        \hline
        \multirow{2}{*}{Learning rate} & $3 \times 10^{3}$ (LL)\\
        & $3 \times 10^{4}$ (MC) \\
        \bottomrule
    \end{tabular}
\end{minipage}
\end{table}

\section{Additional Experiments} \label{S:additional_experiments}
\subsection{Scaling Up the Replay Ratio in Continuous Actions Domains} \label{S:scaling_up_utd_sac}
\begin{figure}[H]
    \centering
    \includegraphics[width=\textwidth]{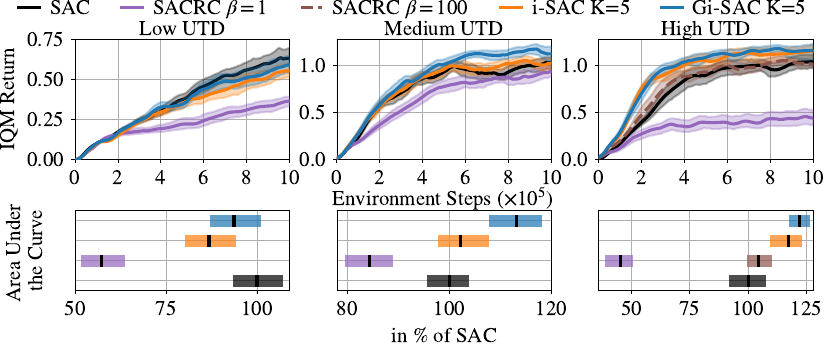}
    \caption{Evaluating the proposed approach on the \textbf{MuJoCo} benchmark with different \textbf{update-to-data~(UTD) ratios} (UTD $\in \{0.1, 1, 10\}$). All algorithms are normalized with respect to SAC for medium UTD to ease comparison. We note that all algorithms diverge for the Humanoids and HumanoidStandUp tasks on the high UTD setting. This is why we aggregate the results over the $4$ remaining MuJoCo tasks (Hopper, Ant, HalfCheetah, and Walker2d). We believe that this divergence is due to the representational capacity of the neural network being too small to cope with high UTD ratios. Importantly, Gi-SAC is competitive against semi-gradient methods, especially for high UTD ratios.}
    \label{F:utd_study_mujoco}
\end{figure}

\clearpage

\subsection{Additional Ablation Studies} \label{S:additional_ablation_studies}
\begin{figure}[H]
    \centering
    \includegraphics[width=\textwidth]{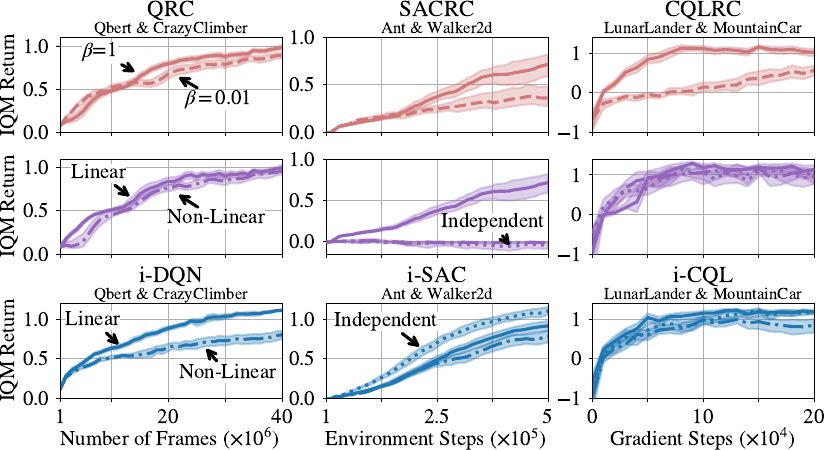}
    \caption{\textbf{Top row}: \textbf{Ablation study} on the \textbf{weight decay coefficient $\beta$} for TDRC. TDRC performs better with a high value of $\beta$ indicating that weight decay is an important component of the algorithm, as suggested in~\citet{ghiassian2020gradient}. \textbf{Center row}: \textbf{Ablation study} on the choice of the \textbf{architecture} for TDRC. Using linear heads on top of a shared feature extractor performs better than the $2$ other considered architectures. \textbf{Bottom row}: \textbf{Ablation study} on the choice of the \textbf{architecture} for i-TD learning. Using linear heads on top of a shared feature extractor performs better than the $2$ other considered architectures, except for the experiment on continuous action-spaces, where the independent architecture performs better. \\
    Overall, we stress that each learning curve is aggregated over $2$ values of UTD ($0.25$ and $1$ for DQN experiments, $1$ and $4$ for SAC experiments) or $2$ dataset sizes for CQL experiments ($10\%$ and $100\%$), and $2$ target update periods ($T=1000$ and $8000$ for DQN experiments, $T=10$ and $1000$ for CQL experiments) or $2$ values of soft-target updates for SAC experiments ($\tau=0.005$ and $0.01$). The learning curves for ablation on the weight decay coefficient are also aggregated over the $3$ archicture designs, except for SACRC, for which we only use the linear shared architecture as the other architecture perform poorly. $\beta$ is set to $1$ for the ablation on the architecture designs.}
    \label{F:tdrc_itd_ablation}
\end{figure}

\clearpage 

\begin{figure}[H]
    \centering
    \includegraphics[width=\textwidth]{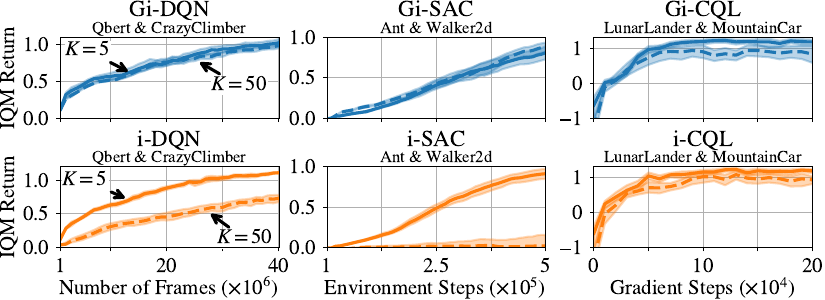}
    \caption{\textbf{Ablation study} on the \textbf{number of Bellman iterations $K$} learned in parallel for i-TD (\textbf{bottom}) and Gi-TD (\textbf{top}) learning. Remarkably, while a performance drop is noticeable when increasing $K$ to $50$ for i-TD learning, Gi-TD performs similarly for the $2$ considered values of $K$. This indicates that Gi-TD learning is less sensitive to this hyperparameter than i-TD learning.}
    \label{F:K_ablation}
\end{figure}

\begin{figure}[H]
    \centering
    \includegraphics[width=\textwidth]{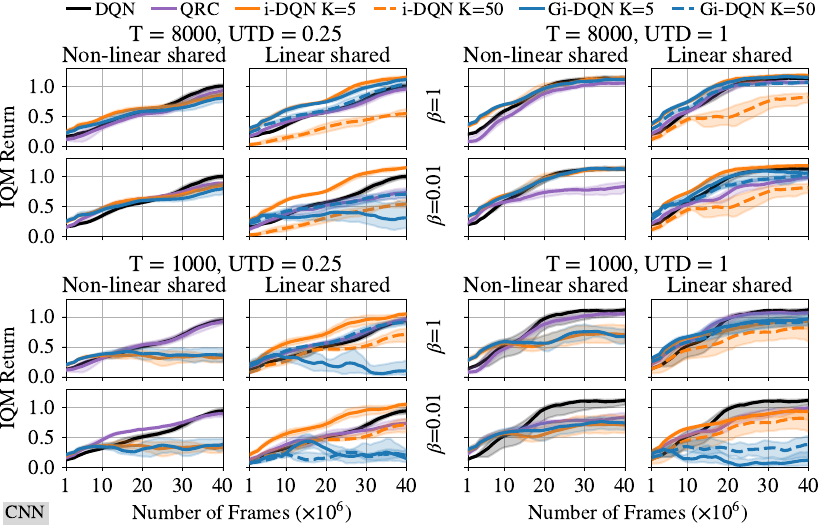}
    \caption{\textbf{Ablation study} over the \textbf{weight decay coefficient $\beta$}, and the choice of the \textbf{architecture} on $2$ \textbf{Atari} games (Qbert, and CrazyClimber). We also vary the target update period $T$, and update-to-data~(UTD) ratio to better grasp the dependency of the algorithms on the studied hyperparameters. DQN with T$=8000$, and UTD$=0.25$ is used to normalize the performance curves.}
    \label{F:dqn_ablation}
\end{figure}

\clearpage 

\begin{figure}[H]
    \centering
    \includegraphics[width=\textwidth]{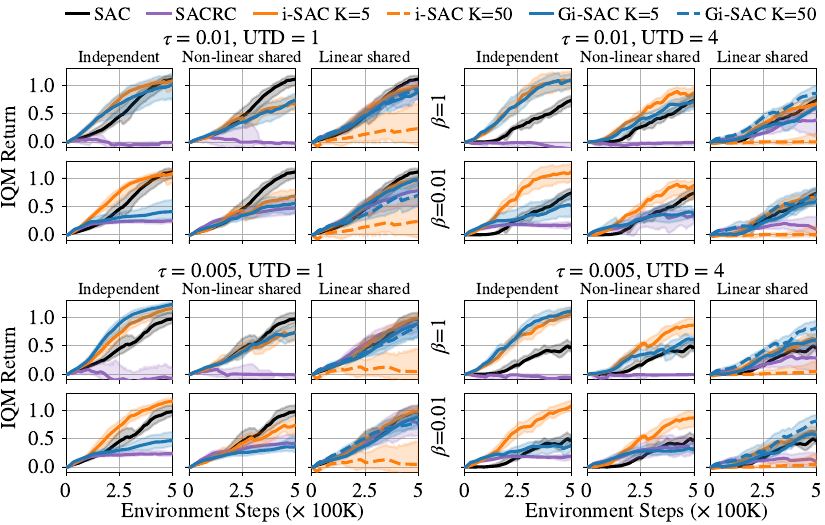}
    \caption{\textbf{Ablation study} over the \textbf{weight decay coefficient $\beta$}, and the choice of the \textbf{architecture} on $2$ \textbf{MuJoCo} tasks (Ant, and Walker2d). We also vary the soft-update coefficient $\tau$, and update-to-data~(UTD) ratio to better grasp the dependency of the algorithms on the studied hyperparameters. SAC with $\tau=0.005$, and UTD$=1$ is used to normalize the performance curves.}
    \label{F:sac_ablation}
\end{figure}

\vspace{-2cm}

\begin{figure}[H]
    \centering
    \includegraphics[width=\textwidth]{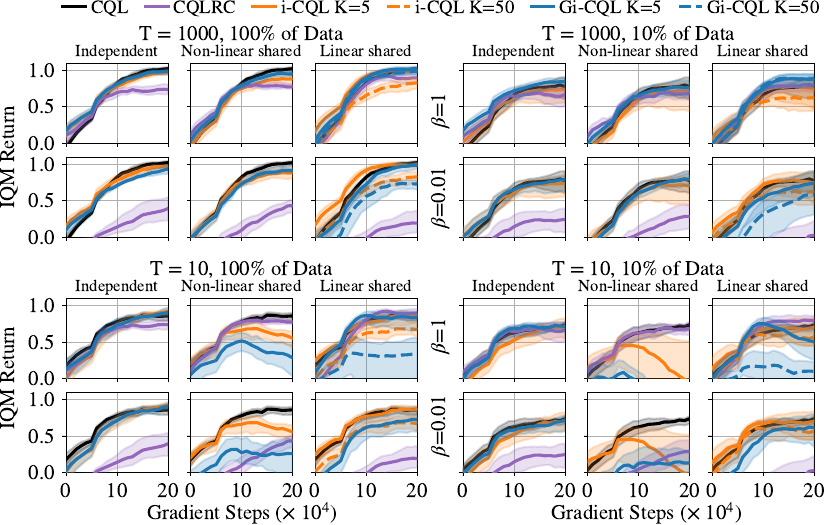}
    \caption{\textbf{Ablation study} over the \textbf{weight decay coefficient $\beta$}, and the choice of the \textbf{architecture} on $2$ \textbf{Gym} environment (LunarLander, and MountainCar). We also vary the target update period $T$, and the percentage of dataset given to the learning algorithm to better grasp the dependency of the algorithms on the studied hyperparameters. The CQL version with T$=1000$ that has access to $100\%$ of the dataset is used to normalize the performance curves.}
    \label{F:cql_ablation}
\end{figure}

\section{Individual Learning Curves} \label{S:individual_learning_curves}
\begin{figure}[H]
    \centering
    \includegraphics[width=\textwidth]{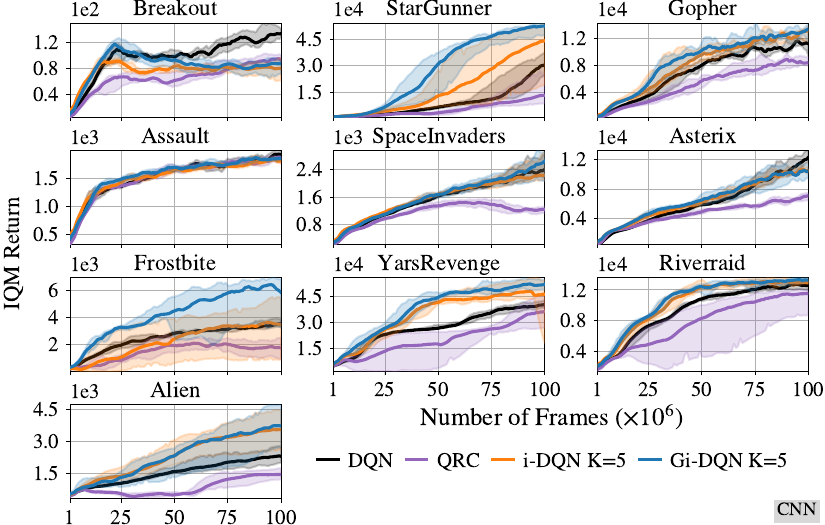}
    \caption{Per \textbf{Atari} game performance for DQN, QRC, i-DQN, and Gi-DQN, in an \textbf{online} learning setting, with the CNN architecture. Except for Breakout, Gi-DQN generally performs better than or is on par with the $3$ other algorithms.
    }
    \label{F:per_env_atari_cnn}
\end{figure}

\begin{figure}[H]
    \centering
    \includegraphics[width=\textwidth]{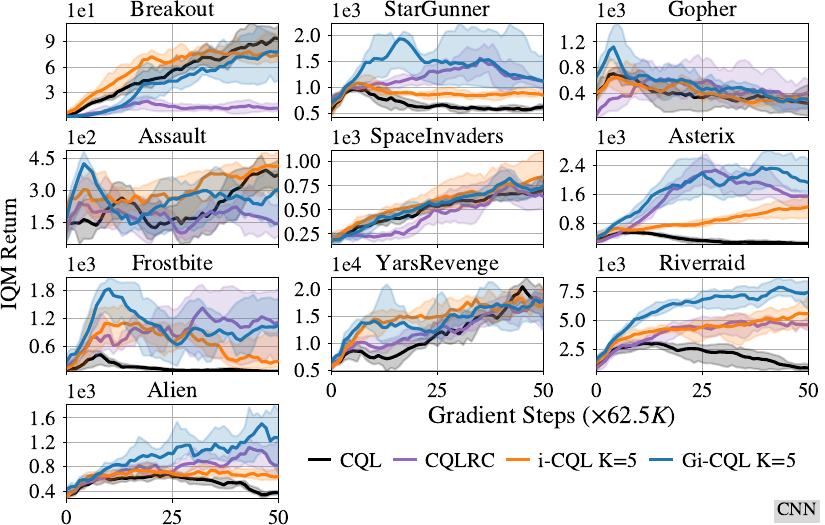}
    \caption{Per \textbf{Atari} game performance for CQL, CQLRC, i-CQL, and Gi-CQL, in an \textbf{offline} learning setting, with the CNN architecture. Except for Breakout and Assault, Gi-CQL generally performs better than or is on par with the $3$ other algorithms.}
    \label{F:per_env_atari_cql}
\end{figure}

\begin{figure}[H]
    \centering
    \includegraphics[width=\textwidth]{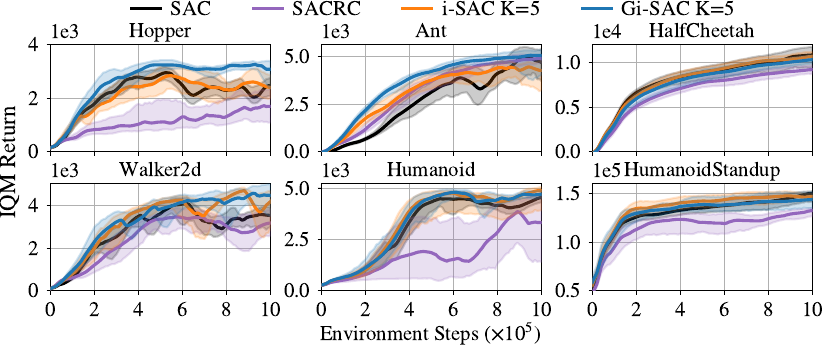}
    \caption{Per \textbf{MuJoCo} task performance for SAC, SACRC, i-SAC, and Gi-SAC, in an \textbf{online} learning setting, with a $2$-layer MLP architecture. Gi-SAC learns more reliably across tasks than SACRC, while remaining at least competitive with i-SAC, if not outperforming it.}
    \label{F:per_env_mujoco}
\end{figure}

\begin{figure}[H]
    \centering
    \includegraphics[width=\textwidth]{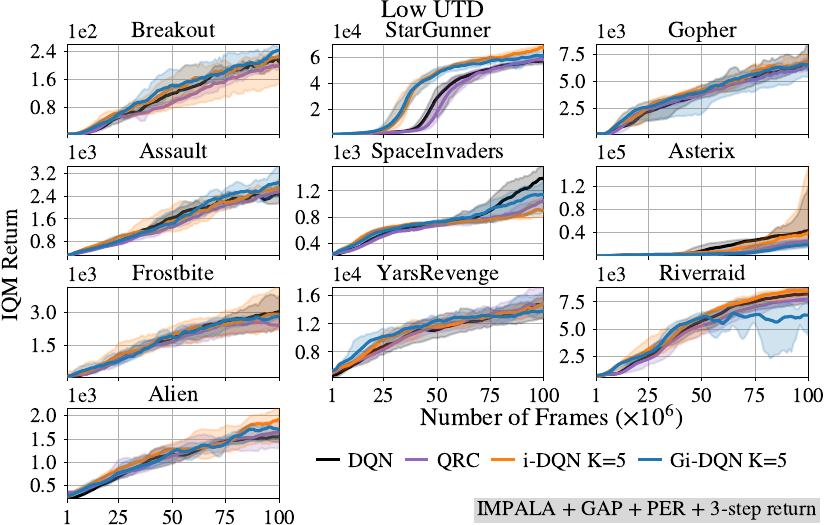}
    \caption{Per \textbf{Atari} game performance for DQN, QRC, i-DQN, and Gi-DQN, in an \textbf{online} learning setting, with the CNN architecture, and a UTD ratio of $\frac{1}{32}$. All algorithms use the IMPALA architecture with global average pooling (GAP), a prioritized experience replay buffer (PER), and a $3$-step return. Due to the low UTD value, all algorithms perform similarly.}
    \label{F:per_env_atari_low_utd}
\end{figure}

\clearpage 

\begin{figure}[H]
    \centering
    \includegraphics[width=\textwidth]{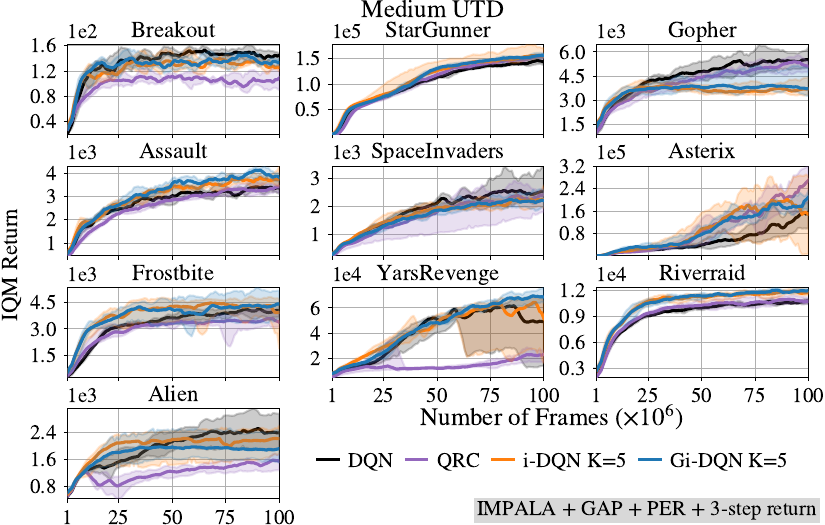}    
    \caption{Per \textbf{Atari} game performance for DQN, QRC, i-DQN, and Gi-DQN, in an \textbf{online} learning setting, with the CNN architecture, and a UTD ratio of $\frac{1}{4}$. Gi-DQN remains competitive against semi-gradient methods even when the baseline algorithm is equipped with the IMPALA architecture with global average pooling (GAP), a prioritized experience replay buffer (PER), and a $3$-step return.}
    \label{F:per_env_atari_medium_utd}
\end{figure}

\vspace{-2cm}

\begin{figure}[H]
    \centering
    \includegraphics[width=\textwidth]{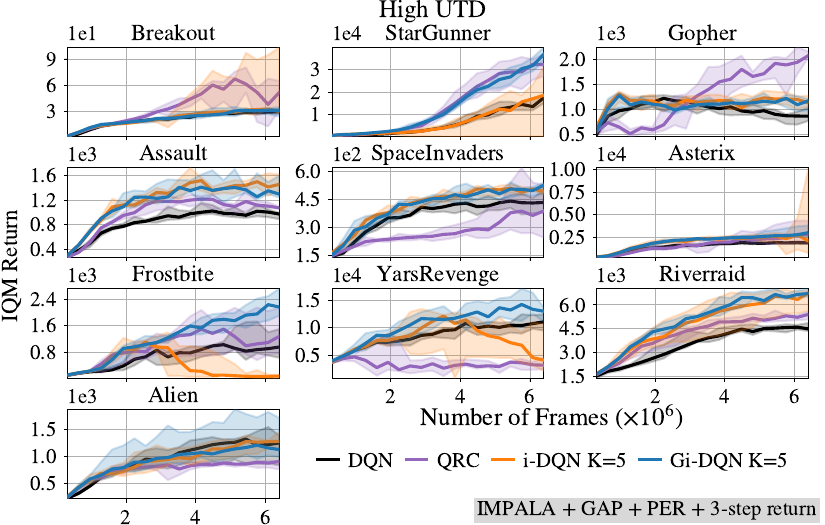}    
    \caption{Per \textbf{Atari} game performance for DQN, QRC, i-DQN, and Gi-DQN, in an \textbf{online} learning setting, with the CNN architecture, and a UTD ratio of $4$. All algorithms use the IMPALA architecture with global average pooling (GAP), a prioritized experience replay buffer (PER), and a $3$-step return. Except for Gopher, Gi-DQN provides a significant boost in learning speed compared to the other algorithms.}
    \label{F:per_env_atari_high_utd}
\end{figure}

\begin{figure}[H]
    \centering
    \includegraphics[width=\textwidth]{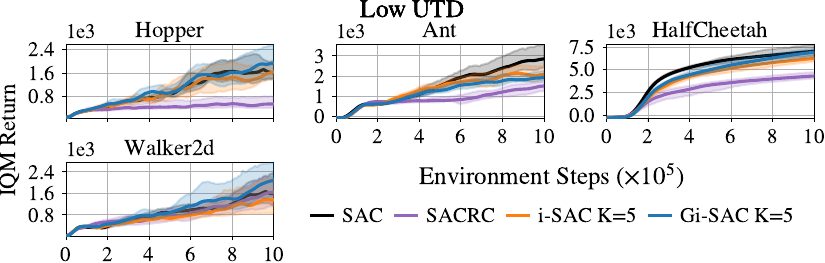}
    \caption{Per \textbf{MuJoCo} task performance for SAC, SACRC, i-SAC, and Gi-SAC, in an \textbf{online} learning setting, with a $2$-layer MLP, and a UTD ratio of $0.1$. SAC, i-SAC, and Gi-SAC reach similar performance.}
    \label{F:per_env_mujoco_low_utd}
\end{figure}

\begin{figure}[H]
    \centering
    \includegraphics[width=\textwidth]{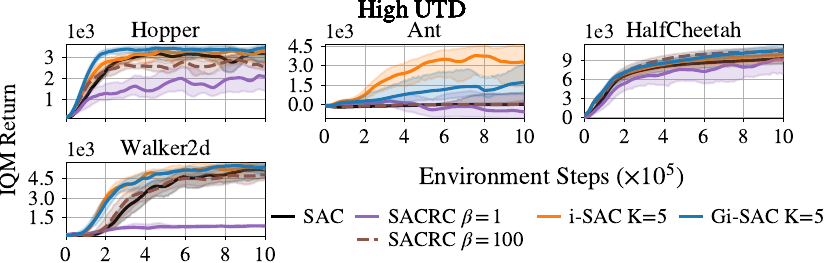}
    \caption{Per \textbf{MuJoCo} task performance for SAC, SACRC, i-SAC, and Gi-SAC, in an \textbf{online} learning setting, with a $2$-layer MLP, and a UTD ratio of $10$. For SACRC, an additional experiment with a higher value of $\beta$ is conducted, since the algorithm underperforms with $\beta=1$. Except for Ant, Gi-SAC performs competitively against i-SAC.}
    \label{F:per_env_mujoco_high_utd}
\end{figure}
\end{document}